\documentclass[sigconf,nonacm]{acmart}
\usepackage{caption}
\usepackage{multirow}
\usepackage{xcolor}
\usepackage{graphicx}
\usepackage{epstopdf}
\usepackage{enumitem}
\setlist{nosep}
\usepackage{url}
\usepackage{comment}
\usepackage{graphicx}
\usepackage{caption}
\usepackage{subcaption}
\AtBeginDocument{%
  \providecommand\BibTeX{{%
  \normalfont 
  \kern-0.5em{\scshape i\kern-0.25em b}\kern-0.8em\TeX}
  }}

\usepackage{amssymb}
\usepackage{amsmath} 
\begin{document}

\title{Can an unsupervised clustering algorithm reproduce a categorization system?}




\author{Nathalia Castellanos}
\email{nathalia.castellanos@blackrock.com}
 \affiliation{%
   \institution{BlackRock, Inc.}
   \city{Atlanta, GA}
   \country{USA}
 }

 \author{Dhruv Desai}
 \email{dhruv.desai1@blackrock.com}
 \affiliation{%
   \institution{BlackRock, Inc.}
   \city{New York, NY}
   \country{USA}
 }
 \author{Sebastian Frank}
 \email{sebastian.frank@blackrock.com}
 \affiliation{%
   \institution{BlackRock, Inc.}
   \city{New York, NY}
   \country{USA}
 }
 \author{Stefano Pasquali}
 \email{stefano.pasquali@blackrock.com}
 \affiliation{%
   \institution{BlackRock, Inc.}
   \city{New York, NY}
   \country{USA}
 }
 \author{Dhagash Mehta}
 \email{dhagash.mehta@blackrock.com}
 \affiliation{%
   \institution{BlackRock, Inc.}
   \city{New York, NY}
   \country{USA}
}

\renewcommand{\shortauthors}{Castellanos et al.}

\begin{abstract}
Peer analysis is a critical component of investment management, often relying on expert-provided categorization systems. These systems' consistency is questioned when they do not align with cohorts from unsupervised clustering algorithms optimized for various metrics. We investigate whether unsupervised clustering can reproduce ground truth classes in a labeled dataset, showing that success depends on feature selection and the chosen distance metric. Using toy datasets and fund categorization as real-world examples we demonstrate that accurately reproducing ground truth classes is challenging. We also highlight the limitations of standard clustering evaluation metrics in identifying the optimal number of clusters relative to the ground truth classes. We then show that if appropriate features are available in the dataset, and a proper distance metric is known (e.g., using a supervised Random Forest-based distance metric learning method), then an unsupervised clustering can indeed reproduce the ground truth classes as distinct clusters.
\end{abstract}

\maketitle

\section{Introduction}
Peer analysis is crucial in finance, grouping entities with shared characteristics to reveal patterns and generate signals with wide-ranging applications starting from risk management, investment strategy, security analysis, relative value analysis, uncovering insights, etc. \cite{damodaran2012investment}. In practice, different classification systems such as Global Industry Classification Standards (GICS) classification \cite{bhojraj2002my}, Morningstar Categorization \cite{morningstarcategorization}, Lipper Categorizations \cite{lippercategory}, various Bond Ratings, etc., are used as a starting point for identifying peers where all the securities or assets within a chosen category or rating are by definition peers of each other. 

However, the consistency of these classification systems often comes into question. Furthermore, they are often misunderstood by comparing and contrasting them with clusters from unsupervised clustering algorithms. This comparison relies on corresponding datasets that capture some or most features that best describe these categorization systems \cite{jung-clustering,elton2003incentive}.

The most notable case study with sizeable literature on the topic is on Morningstar Categorization: Starting from Ref.~\cite{marathe1999categorizing}, studies applied clustering to mutual and hedge funds such as K-means clustering on various fund composition and returns-related variables and argued that Morningstar ``misclassified`` 43\% of the mutual funds available in their dataset as the corresponding Morningstar categories (the ground truth labels) did not uniquely map to the clusters. In Refs. such as \cite{haslem2001morningstar,sakakibara2015clustering,moreno2006self}, funds from various universes such as US large-cap, Japanese funds, Spanish funds, etc., were attempted to be clustered using various variables using different distance metrics and techniques and found to be inconsistent with respect to the respective Morningstar categories.

In a rebuttal to \cite{haslem2001morningstar}, Morningstar researchers argued that it was unreasonable to expect that their categorization system, devised by a committee of experts and relying on both proprietary and public data, should be reproducible simply by clustering on a few variables. Then, in Ref.~\cite{mehta2020machine}, the authors proposed the problem to reproduce Morningstar categories using machine learning (ML) as a supervised classification problem, showing fund categories were indeed learnable with near-perfect accuracy, confirming the categorization system's rule-based consistency. 

However, the challenge of reproducing Morningstar Categorization using purely data-driven unsupervised clustering techniques remained unresolved until Ref. \cite{desai2021robustness}, which first pointed out that although most of the previous research used an appropriate list of variables and evaluation metrics to identify the optimal number of clusters, they were using an inappropriate distance metric (e.g., Euclidean, or other arbitrary ones) for the given dataset and the problem. 
The authors then showed that even a straightforward K-means algorithm using appropriate variables and the distance metric that is learned using Mahalanobis-based distance learning metric learning (DML) method \cite{dml-xing} can indeed reproduce the Morningstar categorization. Hence, the aforementioned problem was solved though only partially as the DML employed in that paper only could deal with numerical variables but not necessarily with categorical or mixed-type variables.

In the present work, we resolve this problem for all numerical, categorical, and mixed-type datasets using a novel distance metric learning method based on Random Forest (RF) proximities. We first use Euclidean distance-based clustering as our baseline, and then apply supervised distance metric learning with Mahalanobis distance \cite{dml-xing}, and lastly, apply a recent technique using Random Forest proximities known as RF-based Potential of Heat-diffusion for Affinity-based Transition Embedding (RF-PHATE) \cite{rhodes2023geometry, moon2019visualizing,rf-phate-rhodes}.  
\vspace{-4mm}
\subsection{Previous Works and Our Contribution}
Clustering algorithms and their evaluation metrics are extensively reviewed in Refs.~\cite{ezugwu2022comprehensive, james2013introduction}. Various Refs~\cite {kapil2016performance, kumar2014performance} have investigated K-means clustering with different distance metrics and concluded that the performance of the algorithm indeed varied with different distance metrics and for different data. However, in these references, the authors used pre-defined distance metrics rather than learning the distance metric from data.

Supervised clustering, where distance metric is learned with the help of ground truth data and then fed into a supervised algorithm has been extensively studied (see, e.g., \cite{aggarwal1999merits,de2020metric,suarez2021tutorial}) although not in the context of directly investigating cluster qualities using internal and external evaluation metrics.

Refs.\cite{ma2023need, han2022adbench} investigated a similar question on the effectiveness of unsupervised learning on the effectiveness of unsupervised learning algorithms and the appropriateness of their evaluation metrics in the context of anomaly detection. In Ref.~\cite{ma2023need,han2022adbench}, the authors performed an extensive review of the existing literature and experiments on various internal evaluation metrics for unsupervised anomaly detection models to conclude that none of these evaluation metrics were practically helpful in identifying the best model.

Our overall approach can be summarized as follows: for a given labeled dataset, we first show that when the target variable is masked, merely using K-means with Euclidean distance and using either internal or external evaluation metrics to identify the optimal number of clusters ($K_C$) usually does not reproduce ground truth classes as distinct clusters (by which we mean that each cluster has data-points from one and only one ground truth class). However, with the help of the target variable, if we learn a distance metric for the dataset in a supervised fashion using the Mahalanobis \cite{dml-xing, desai2021robustness} and RF-based \cite{rhodes2023geometry, rf-phate-rhodes} distance metric learning methods, then the new distance metrics with K-means can yield much better results in reproducing ground truth classes as distinct clusters.

We also show that the success of reproducing the ground truth classes using clustering depends on whether the available input features are highly predictive of the target variable by analyzing the correlation between the accuracy of RF for the classification task and internal and external evaluation metrics for clustering.

\section{Methodology}\label{sec:methodology}
In this Section, we describe the methodology and techniques used in the present work.
\subsection{K-means Clustering:} K-means clustering \cite{james2013introduction} is a widely-used unsupervised learning algorithm that partitions a dataset of size $n$ into a predefined number (K) of compact and well-separated clusters by minimizing the within-cluster variances. Each cluster $C_k$ is defined by its centroid $\mu_k$, for $i = 1, \dots, K$. The objective function of K-means is to minimize the within-cluster sum of 
squares:
\begin{equation*}\label{eq:l2}
\sum_{k=1}^{K} \sum_{x \in C_k} \|x - \mu_k\|^2,
\end{equation*} 
where $x$ represents a data point in the dataset. 
Essential to any clustering problem is the selection of an appropriate distance metric. A valid distance measure for instances $x$, $y$, and $z$ in $\mathbb{R}^{m}$ must follow the following axioms: (1) $d(x,y) \geq 0$, (2) $d(x,y) = 0$ if and only if $x = y$, (3) ${d(x,y)} = d(y,x)$, and (4) $d(x,y) + d(y,z) \leq d(x,z)$ for all $x, y, z \in \mathbb{R}^m$.

We perform K-means clustering using the following distance metrics: Euclidean distance, Mahalanobis distance metric (learned in a supervised fashion), RF-based distance metric. The latter two are pseudo-metrics, i.e., they do not necessarily satisfy property (2). For each of the chosen distance metrics, we run K-means clustering for K ranging from 3 to 100.

\subsection{Supervised Clustering using Mahalanobis Metric}
While Euclidean distance, defined as ${d(x,y) = \sqrt{\sum^{n}_{i=1}(x_{i}-y_{i})^{2}}}$, is useful when dealing with vectors of the same scale, it can be inadequate when measuring the distance across multiple dimensions. Instead, the Mahalanobis metric transforms disparate numeric features into a scale-invariant space and is defined as \cite{mahalanobis2018generalized} ${d}_{M}(x,y) = \sqrt{(x - {y})^{T}M(x - y)}$, where $M$ is the inverse matrix of weights, over which points are normalized. When $M$ is the identity matrix, $d_M(x,y)$ is the Euclidean distance. Note that the Mahalanobis distance may only be applied to numeric variables.

In Ref.~\cite{dml-xing}, a distance metric learning method based on the Mahalanobis metric was posed as a convex optimization problem to find the values of the entries of $M$ such that they minimize the distances between the most similar points (those from the same class, for example) while maximizing the distances between those which are dissimilar (e.g., pairs from two different classes). More specifically, the following optimization problem is solved to learn the distance metric from the given dataset: 
\begin{align*}
    & \underset{M}{min}\sum^{}_{(x_{i},x_{j})\in\mathbb{S}}d_{M}(x_{i},x_{j}), \\
    & {s.t}\sum^{}_{(x_{k},x_{l})\in\mathbb{D}}d^{2}_{M}(x_{k},x_{l})\geq1,\quad and \quad {M}\succcurlyeq{0}.
\end{align*}
Here, ${M}\succcurlyeq{0}$ means that $M$ should be positive-definite. Once the final $M$ is obtained, one can rescale the input features as $\sqrt{M}x \rightarrow x_{MMC}$ and use K-means algorithm as defined above but now using $x_{MMC}$ as the coordinates \cite{desai2021robustness}. We call this Mahalanobis metric based distance metric learning as Mahalanobis Metric (MMC) learning.

\subsection{Supervised Clustering using RF-PHATE}
The random forest (RF) \cite{breiman2001random} is a popular ensemble-based supervised ML model that requires minimal preprocessing of the data, handles missing values as well as fairly large datasets, and is robust to overfitting. The RF is equivalent to an adaptive weighted k-nearest-neighbor algorithm that learns a local distance metric from the given data in a supervised fashion. However, extracting the actual learned distance metric from the trained RF posed a challenge until recently when an explicit formula was derived to compute pairwise similarities, based on what the authors designate as the Geometry and Accuracy Preserving (GAP) proximity \cite{rhodes2023geometry} and is defined for a trained RF as:
\begin{equation} \label{GAP}
k^{GAP}_{i,j} = \frac{1}{|S_{i}|} \sum_{t \in S_{i}} \frac{c_j(t)I[j \in J_{i}(t)]}{|M_{i}(t)|},
\end{equation}
where $S_{i}$ is the set of trees for which observation $i$ is out of bag. $M_i(t)$ is the multiset (i.e., a set that allows for repetition among the elements of the set) of bagged points in the same leaf as $i$ in tree $t$, $J_{i}(t)$ is the corresponding set (i.e., without repetitions) of bagged points in the same leaf as $i$ in tree $t$. Moreover, $c_{j}(t)$ is the multiplicity of the index $j$ in the bootstrap sample.

In recent work, the GAP proximity has been used along with a technique called Potential of Heat-diffusion for Affinity-based Transition Embedding (PHATE) \cite{rf-phate-rhodes} which is a dimensionality reduction technique designed to capture both local and global structures of a given dataset. It uses a diffusion-based approach to compute low-dimensional embeddings that preserve the intrinsic geometry of the data, making it suitable for visualizing complex datasets.
\begin{table*}[htbp]
\centering
\begin{tabular}{l l p{3.55in} l} 
 \toprule
 Metric Name & Metric Type & Description & Range \\ \midrule
 Inertia* & Internal & Sum of squared distances from each point to its assigned cluster centroid &  \(\geq 0\)  \\  
 Silhouette Score \cite{rousseeuw1987silhouettes} & Internal & Measures how similar an object is to its own cluster compared to other clusters & \([-1, 1]\)  \\ 
 Calinski-Harabasz Index \cite{calinski1974dendrite} & Internal & Ratio of between-cluster dispersion to within-cluster dispersion & \(\geq 0\)  \\ 
 Davies-Bouldin Index* \cite{davies1979cluster} & Internal & Average similarity ratio of each cluster with its most similar cluster &  \(\geq 0\)  \\ 
 Gap Statistics \cite{tibshirani2001estimating} & Internal & Compares within-cluster dispersion to expected dispersion under a null reference distribution &  \(\geq 0\)  \\ 
 Homogeneity \cite{v-measure-rosenberg} & External & All clusters contain only members of a single class & \([0, 1]\)  \\ 
 Completeness \cite{v-measure-rosenberg} & External & All members of a given class are assigned to the same cluster & \([0, 1]\)  \\ 
 V-measure \cite{v-measure-rosenberg} & External & Harmonic mean of homogeneity and completeness & \([0, 1]\)  \\ 
 Rand Index \cite{rand-index} & External & Similarity of the assignment of pairs of elements in clusters & \([0, 1]\)  \\ 
 Adjusted Rand Index \cite{rand-index} & External & Rand Index adjusted for chance & \([-1, 1]\)   \\ 
 Normalized Mutual Information \cite{vinh2010information} & External & Information shared between clusters, normalized & \([0, 1]\)  \\ 
 Fowlkes-Mallows Score \cite{fowlkes1983method} & External & Geometric mean of precision and recall & \([0, 1]\)  \\  \bottomrule
\end{tabular}
\caption{Summary table for evaluation metrics for clustering algorithms. Metrics marked with an asterisk (*) indicate that lower values denote better performance.}
\label{tab:eval_metrics_summary}
\vspace{-4mm}
\end{table*}
The RF-based PHATE, or RF-PHATE, algorithm follows the following steps: (1) train an RF on the given dataset to predict labels and generate GAP proximities; (2) learn the local structure of the relevant variables using GAP proximities; (3) learn the global structure of the relevant variables by diffusing the GAP proximities; (4) extract the local and global structure via the potential distances.

\subsubsection{Multidimensional Scaling (MDS)} MDS is a powerful technique to visualize the inherent structure of high-dimensional data by representing it in a lower-dimensional space while preserving its pairwise distances as much as possible. MDS plays a crucial role in the final step of the RF-PHATE algorithm.

RF-PHATE combines RF proximities with the dimensionality reduction method PHATE, leveraging Multidimensional Scaling (MDS) to visualize the intrinsic structure of the data relevant to the supervised task. In the present work, our main goal is not to visualize the data but to employ the distance metric learned by RF through RF-PHATE within the context of K-means clustering. 

Hence, instead of taking only two components of MDS, we determine the optimal number of dimensions for embedding high-dimensional data as follows: after training the RF to predict the ground truth labels, a diffusion operator is constructed to balance local and global structural information by simulating random walks using RF-GAP proximities. Then, the potential distances, derived from the log-transformed diffused probabilities, encode both local and global relationships within the data. MDS is then applied to these potential distances to reduce the dimensionality of the data. The goal is to find the optimal number of dimensions where the MDS objective function (typically referred to as stress or strain) is minimized, indicating the best representation of the data's structure.

This optimal number of dimensions is identified by iterating over a range of possible dimensions (e.g., 2 to 10) and computing the stress value for each. The stress value quantifies how well the lower-dimensional representation preserves the original pairwise distances. The dimensionality corresponding to the lowest stress value is selected as the optimal number of dimensions. Finally, we run the K-means clustering in the $m$-dimensional space.

\section{Evaluation Metrics}
\subsection{Evaluating Random Forest Training}
To evaluate how well an RF is trained for a given classification task, we use weighted F1-score and accuracy (the proportion of correctly classified instances out of the total instances).
Initially, the RF pipeline performs stratified K-Fold cross-validation to evaluate the model’s baseline performance and for hyperparameter (the number of estimators and maximum depth of the trees) tuning with grid-search. The model is then refitted using parameters with the best cross validation performance, and the performance is then re-evaluated on the test set. This final model is used for all the downstream tasks afterward.
\subsubsection{Evaluating K-means Clustering} 
Evaluating unsupervised clustering algorithms is challenging as by definition there is no ground truth label. In the literature, numerous metrics have been proposed to evaluate unsupervised clustering, which can be classified into two broad types \cite{james2013introduction}: internal evaluation metrics which evaluate the quality of clustering without relying on any ground truth labels; and, external evaluation metrics which evaluate the clusters with respect to a set of ground truth labels (though the labels were not taken into consideration when performing the clustering). We use both types of evaluation metrics as outlined in Table \ref{tab:eval_metrics_summary}.

We determine the optimal number of clusters ($K_C$) for each dataset for each of the three distance metrics under investigation by identifying the value of $K$ that optimizes individual evaluation metrics from Table \ref{tab:eval_metrics_summary}.

\subsection{Evaluating How Well Clustering Reproduces Ground Truth Classes}
Clustering accuracy \cite{dml-xing} is a metric used to evaluate the performance of a clustering algorithm by comparing the predicted cluster assignments with a set of true or desired cluster assignments. It is defined as the probability that a randomly selected pair of points is correctly identified as either belonging to the same cluster or different clusters by the clustering algorithm, i.e.,
\begin{equation}
\text{Accuracy} = \sum_{i > j} \frac{1\{1\{c_i = c_j\} = 1\{\hat{c}_i = \hat{c}_j\}\}}{0.5m(m - 1)},
\label{eq:accuracy}
\end{equation}
where $m$ is the total number of data points, $c_i$ and $c_j$ are the true cluster assignments for points $x_i$ and $x_j$, respectively. $\hat{c}_i$ and $\hat{c}_j$ are the cluster assignments predicted by the clustering algorithm for points $x_i$ and $x_j$, respectively. $1_{\{\cdot\}}$ is 1 if the condition inside is true, and 0 otherwise.

\section{Data Description}\label{sec:dataset}
We experimented with an extensive list of public datasets (though here the results are shown for a shorter list of representative datasets for the sake of brevity) arising from diverse areas from the UCI Machine Learning Repository \cite{kelly:2019} as listed in Table \ref{tab:my_label}. 

\begin{table}[]
\centering
\begin{tabular}{ l  l  l  l  l } 
\toprule
Dataset & \multirow{2}{*}{\begin{tabular}[c]{@{}l@{}}n.\\ Instance\end{tabular}} & \multirow{2}{*}{\begin{tabular}[c]{@{}l@{}}n.\\ Feature\end{tabular}} & \multirow{2}{*}{\begin{tabular}[c]{@{}l@{}}n. cat.\\ Feature\end{tabular}} & \multirow{2}{*}{\begin{tabular}[c]{@{}l@{}}n.\\ Class\end{tabular}} \\ 
 & & & & \\ \midrule 
Iris & 150 & 4 & 0 & 3 \\ 
Wine & 178 & 13 & 0 & 3 \\ 
Breast Cancer & 569 & 30 & 0 & 2 \\ 
Ionosphere & 351 & 34 & 0 & 2 \\ 
Balance Scale & 625 & 4 & 0 & 3 \\ 
Algerian Fires & 244 & 10 & 0 & 2 \\ 
Banknote & 1,372 & 4 & 0 & 2 \\ 
Cervical Cancer & 858 & 18 & 0 & 2 \\ 
Car Evaluation & 1,728 & 6 & 6 & 4 \\ 
Mushroom & 8,124 & 22 & 22 & 2 \\ 
\bottomrule
\end{tabular}
\caption{Toy Dataset Description \cite{kelly:2019}}
\label{tab:my_label}
\vspace{-6mm}
\end{table}
\raggedbottom

As a real-world dataset, we used funds data from the Morningstar\footnote{\label{mstar_cright} \copyright 2024 Morningstar. All Rights Reserved. The information contained herein: (1) is proprietary to Morningstar and/or its content providers; (2) may not be copied or distributed; and (3) is not warranted to be accurate, complete, or timely. Neither Morningstar nor its content providers are responsible for any damages or losses arising from any use of this information. Past performance is no guarantee of future results.} Data-warehouse feed \cite{morningstarcategorization}. Morningstar Categories are created based on a fund's portfolio composition. We sourced features that correspond to a fund's portfolio composition across various aggregations. The dataset includes distinct features representing a particular fund’s asset allocation percentages. We used all U.S. domiciled open-end mutual funds and ETFs and subset our data for each fund to a single share class as portfolio aggregation for different share classes would be identical. In the end, we have around 10.5K funds spread across 124 Morningstar categories with highly imbalanced distribution, with 14 numerical and 2 categorical variables, for the March 2024 snapshot. We used one-hot encoding for the categorical variables and interpreted missing values as 0\%. Further details of the data can be found in Ref.~\cite{desai2023quantifying}.

\subsection{Data Preparation}
For K-means clustering using Euclidean distance, we have preprocessed each of the numerical columns by performing standard scaling on each of the columns. For Mahalanobis and RF-based distance metric learning, we have not done any standardization on the numerical columns. We one-hot encoded all the categorical columns before employing any of the algorithms. There were no missing values in the toy datasets. For the fund's data, since the features describe the percentage of funds portfolio allocation in the respective segment, we imputed the missing values with zero.

\section{Computational Details}
We used Sci-kit Learn \cite{scikit-learn} for our K-means clustering, RF training, and most of the evaluation metrics, and coded the remaining ones in Python on our own. For MMC learning, we used the metric-learn\footnote{https://contrib.scikit-learn.org/metric-learn/} package. For RF-PHATE, we used the rfphate\footnote{https://github.com/jakerhodes/RF-PHATE} package, which provided an embedding that we subsequently used for K-means clustering.

\section{Results}\label{sec:results}

\subsection{RF Performance for the Classification Tasks}
First, to show the performance of RF on each of the datasets, we record the weighted F1-score and accuracy of the RF for the best hyperparameter-point for the respective datasets in Table \ref{tab:dataset_rf_comparison}.

\begin{table}[h!]
\centering
\begin{tabular}{lp{1in}l}
\toprule
\textbf{Dataset} & \textbf{Weighted F1 Score} & \textbf{Accuracy} \\ \midrule
Iris & 95.9 & 96.0 \\ 
Wine & 98.8 & 98.9 \\ 
Breast Cancer & 94.3 & 94.4 \\ 
Ionosphere & 93.7 & 93.7 \\ 
Balance Scale & 84.5 & 86.1 \\ 
Algerian Fires & 97.5 & 97.5 \\ 
Banknote & 99.3 & 99.3 \\ 
Cervical Cancer & 88.9 & 90.2 \\ 
Car Evaluation & 98.0 & 97.9 \\ 
Mushroom & 100 & 100 \\ 
Funds & 94.6 & 94.8 \\\bottomrule
\end{tabular}
\caption{RF performances on toy datasets for the best hyperparameter-points on complete datasets.}
\label{tab:dataset_rf_comparison}
\vspace{-6mm}
\end{table}

\subsection{Internal Evaluation Metrics}
Next, we evaluate K-means clustering with various distance metrics using internal evaluation metrics:

\subsubsection{Standard K-means Clustering (Euclidean Distance)}

\begin{figure*}[htbp]
    \centering

    \begin{subfigure}[t]{0.2\textwidth}
        \includegraphics[width=\textwidth]{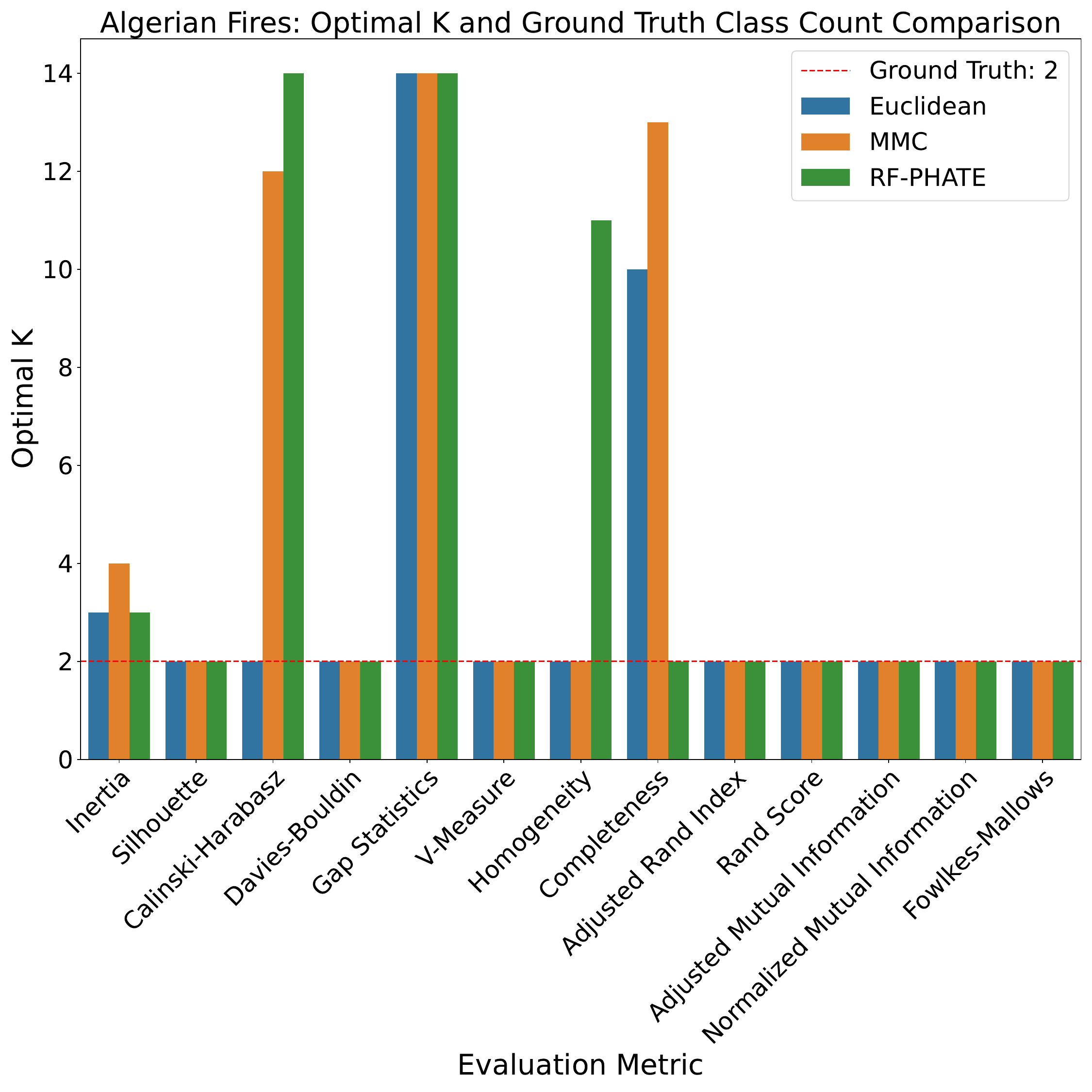}
    \end{subfigure}
    \hfill
    \begin{subfigure}[t]{0.2\textwidth}
        \includegraphics[width=\textwidth]{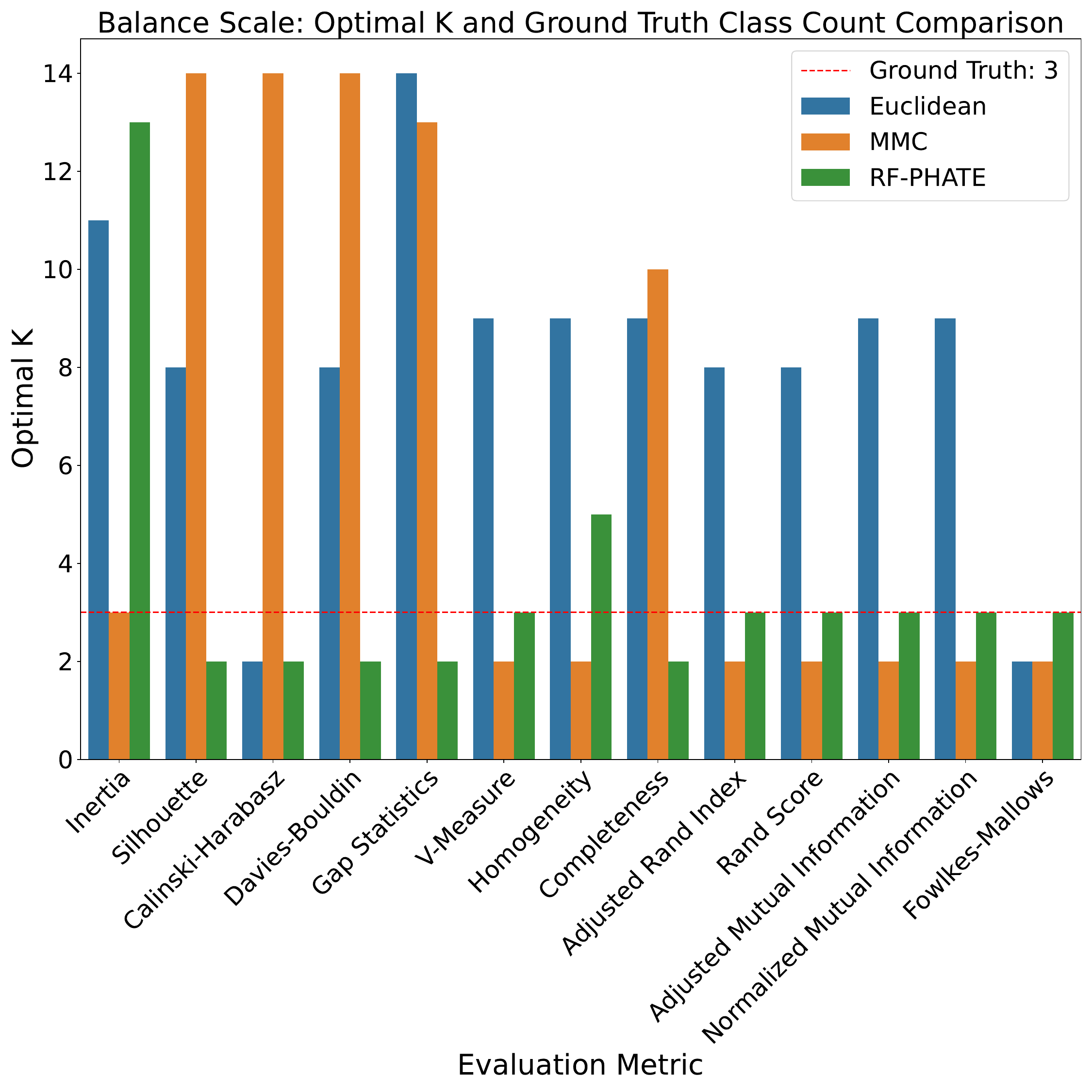}
    \end{subfigure}
    \hfill
    \begin{subfigure}[t]{0.2\textwidth}
        \includegraphics[width=\textwidth]{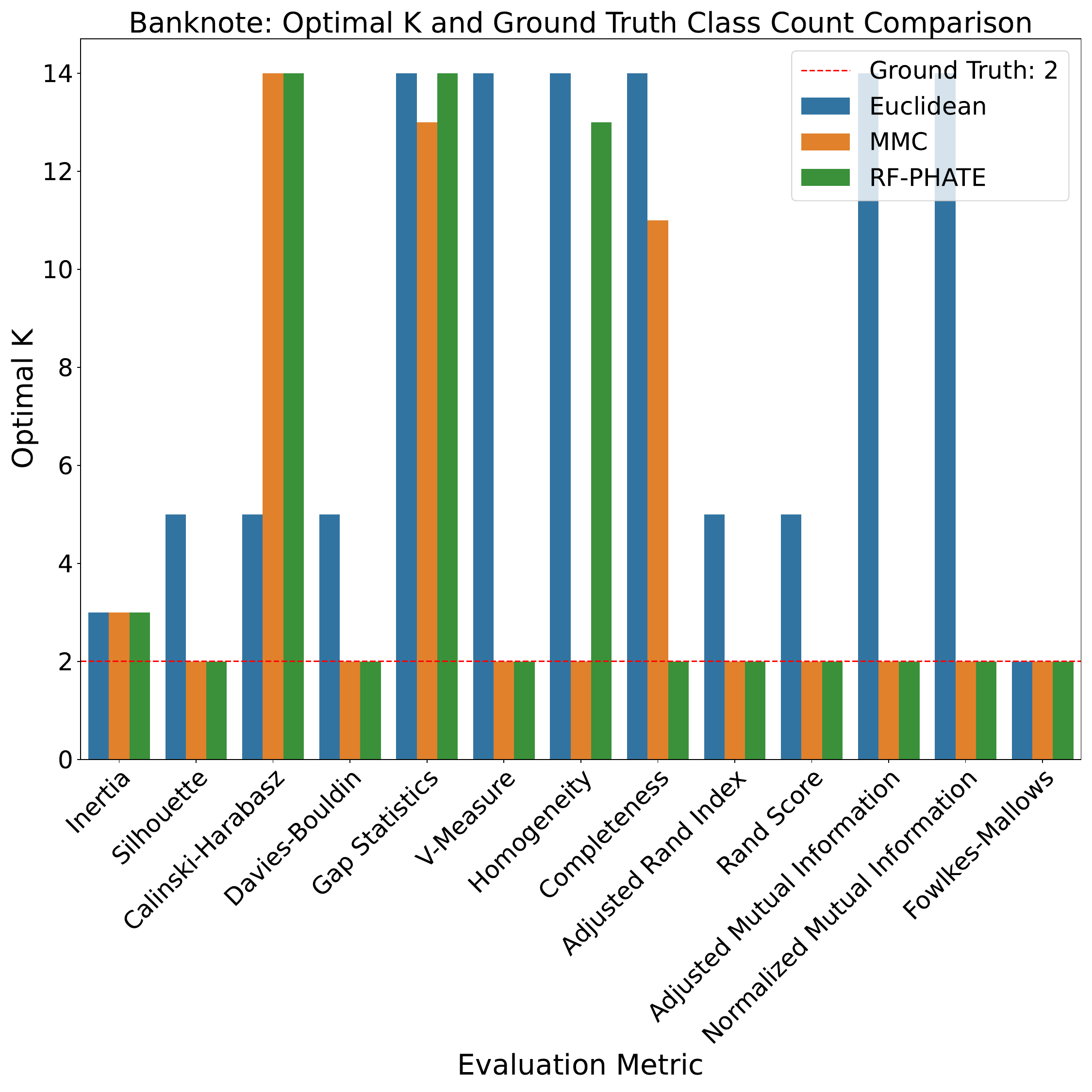}
    \end{subfigure}
    \hfill    
    \begin{subfigure}[t]{0.2\textwidth}
        \includegraphics[width=\textwidth]{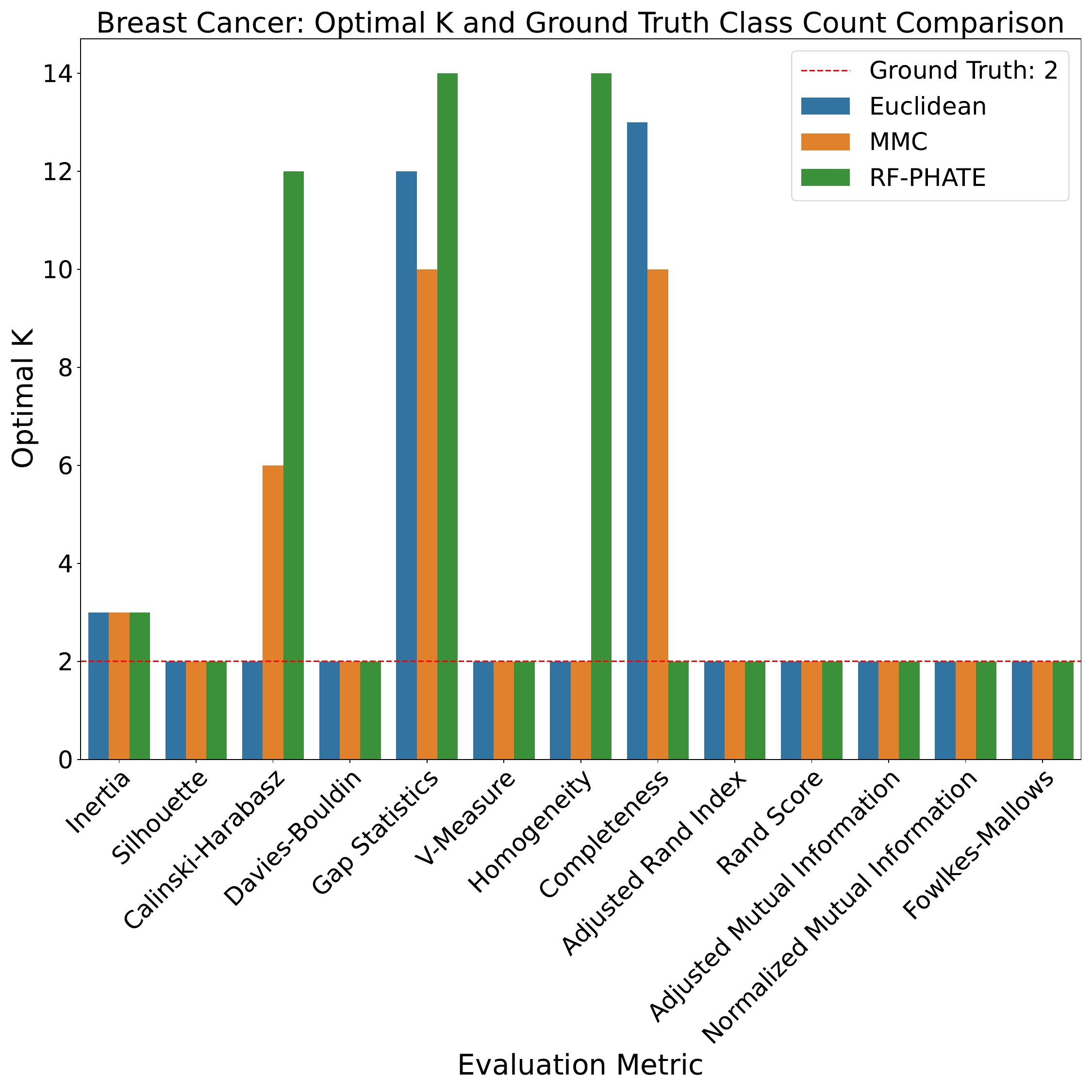}
    \end{subfigure}
    
    \begin{subfigure}[t]{0.2\textwidth}
        \includegraphics[width=\textwidth]{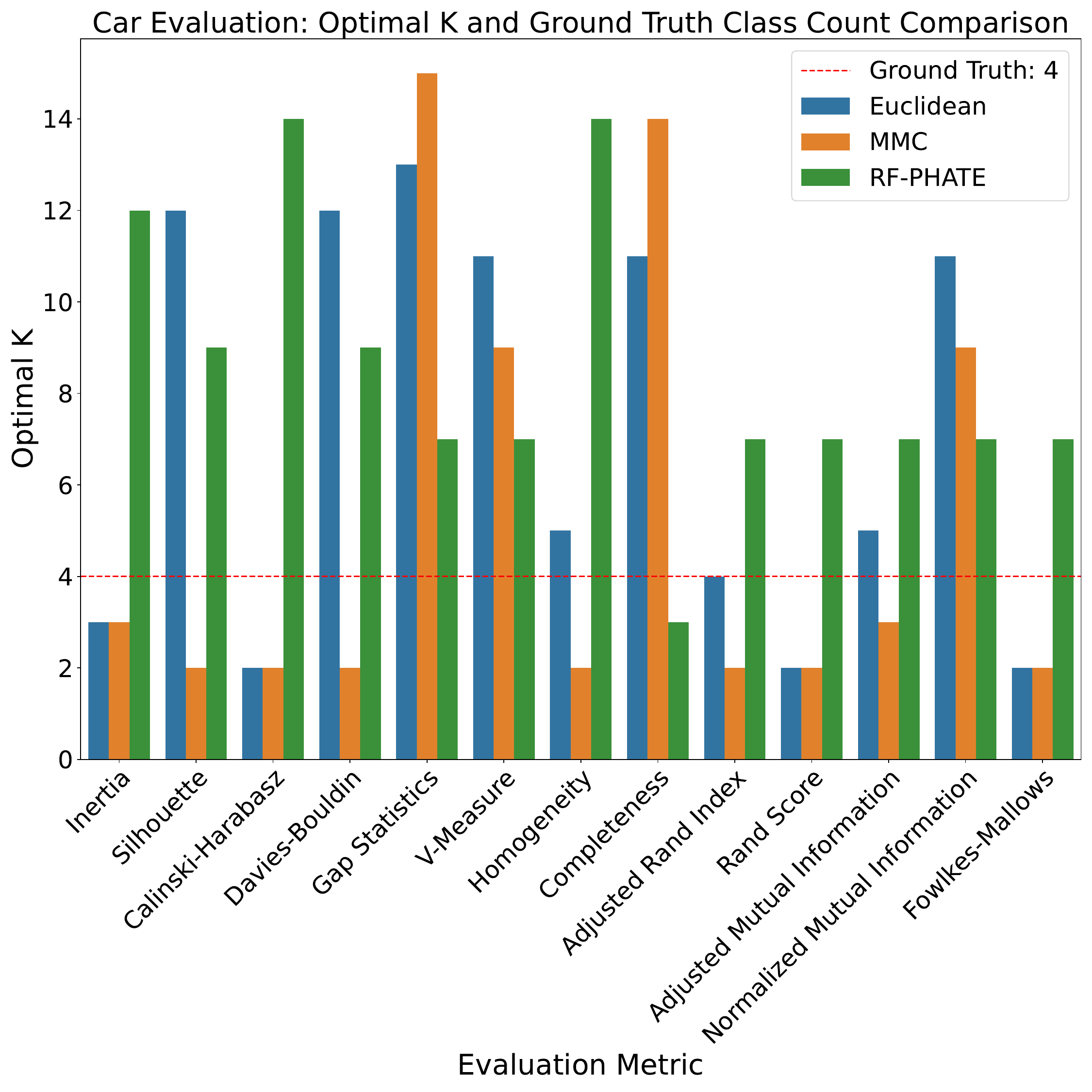}
    \end{subfigure}
    \hfill
    \begin{subfigure}[t]{0.2\textwidth}
        \includegraphics[width=\textwidth]{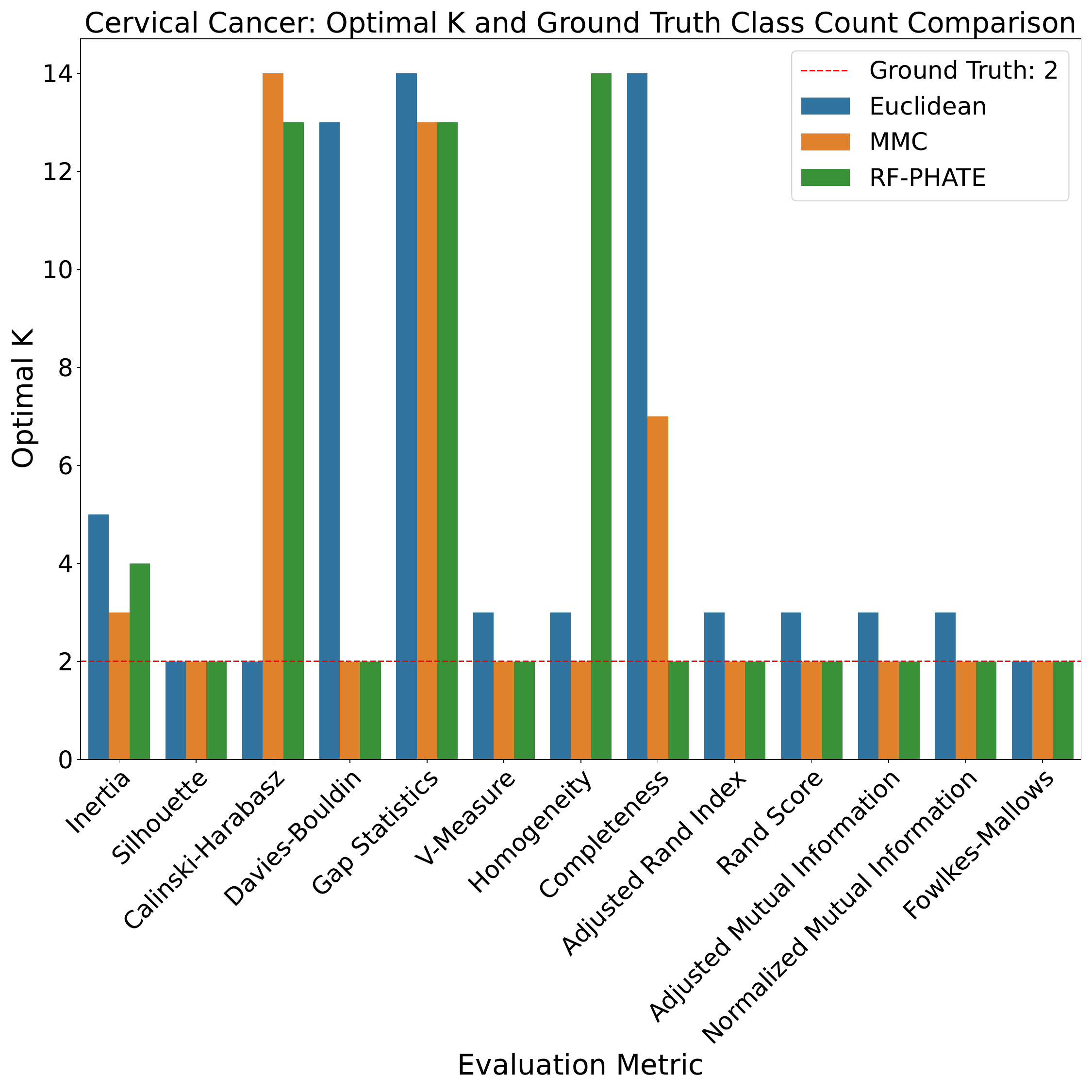}
    \end{subfigure}
    \hfill
    \begin{subfigure}[t]{0.2\textwidth}
        \includegraphics[width=\textwidth]{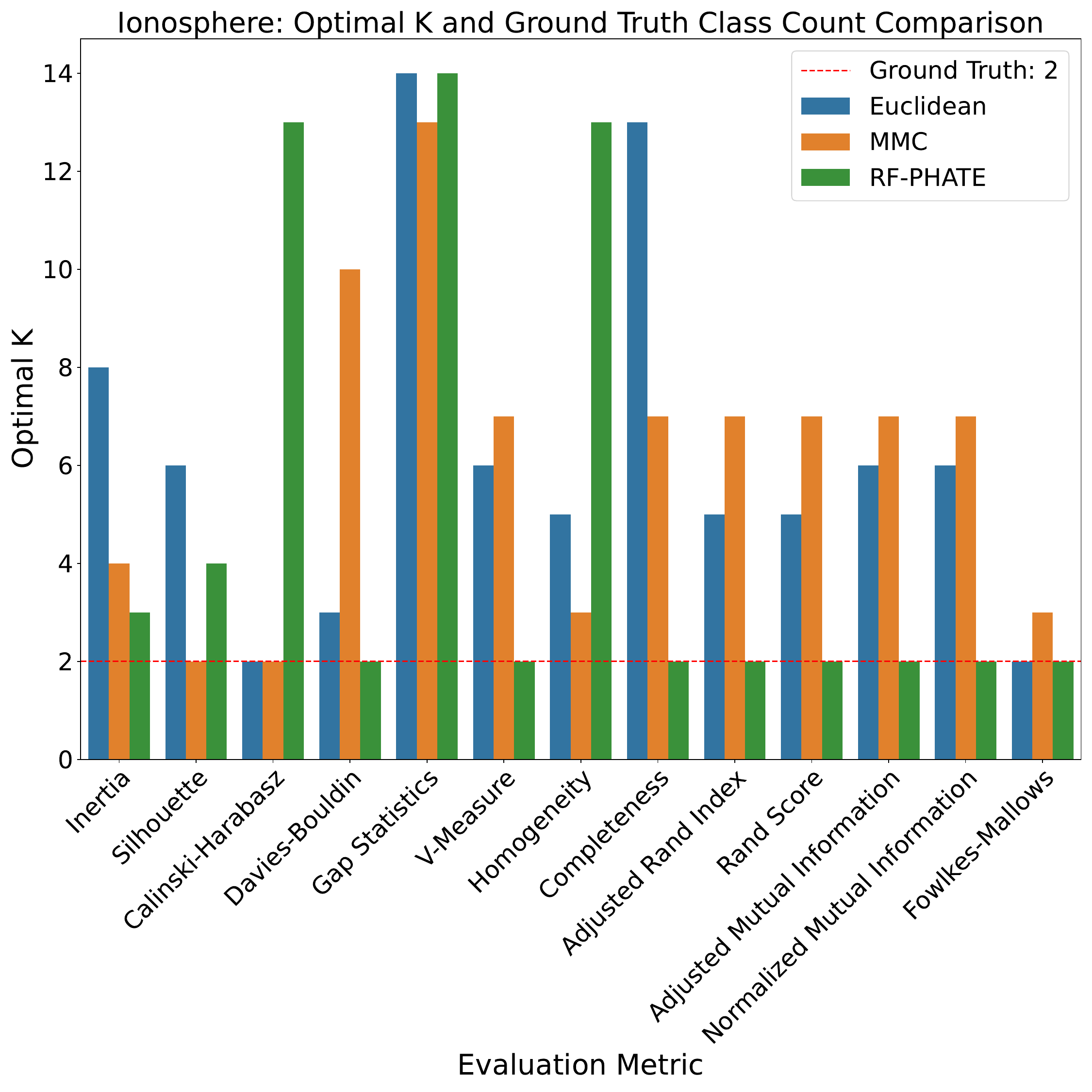} 
    \end{subfigure}
    \hfill
    \begin{subfigure}[t]{0.2\textwidth}
        \includegraphics[width=\textwidth]{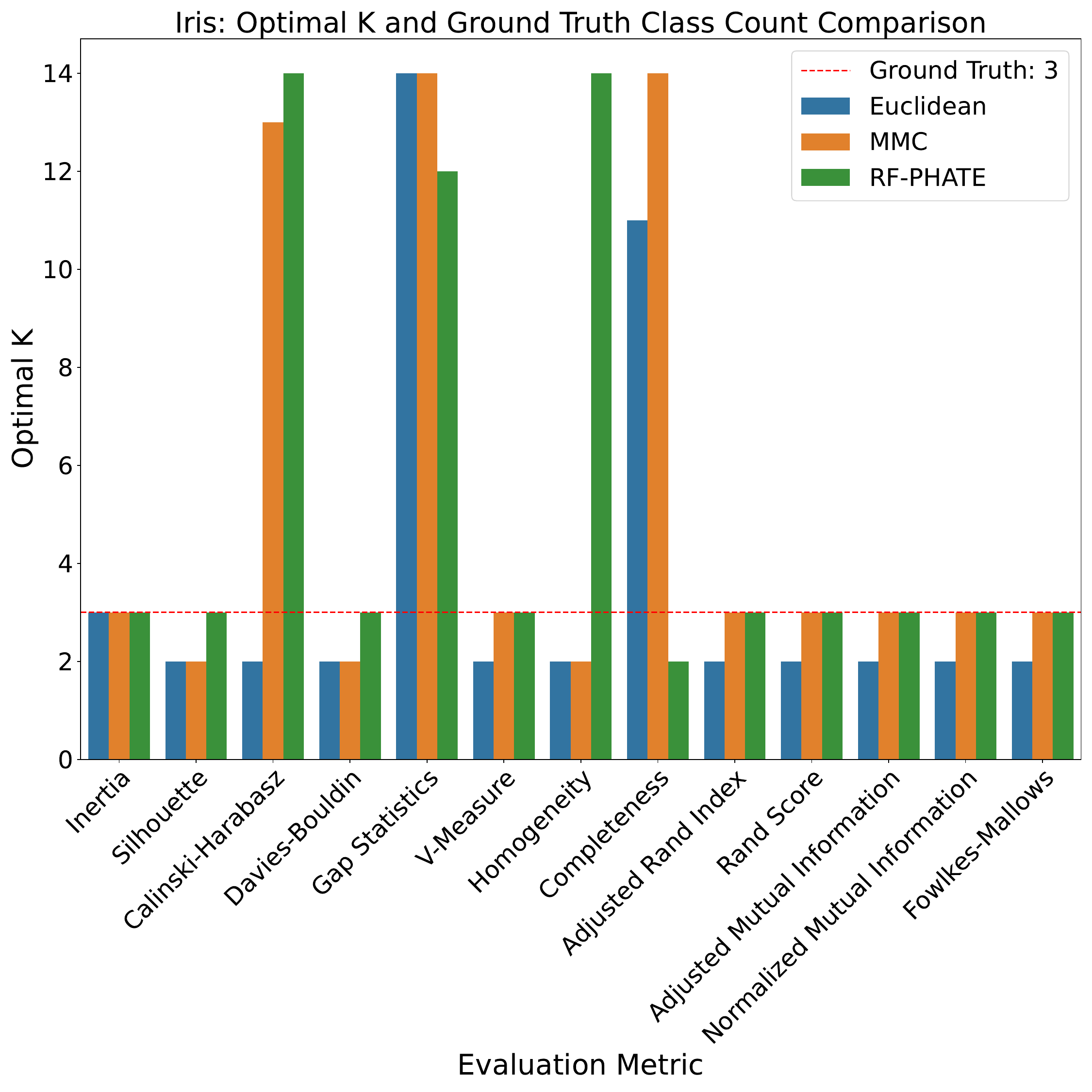}
    \end{subfigure}
    
    \begin{subfigure}[t]{0.2\textwidth}
        \includegraphics[width=\textwidth]{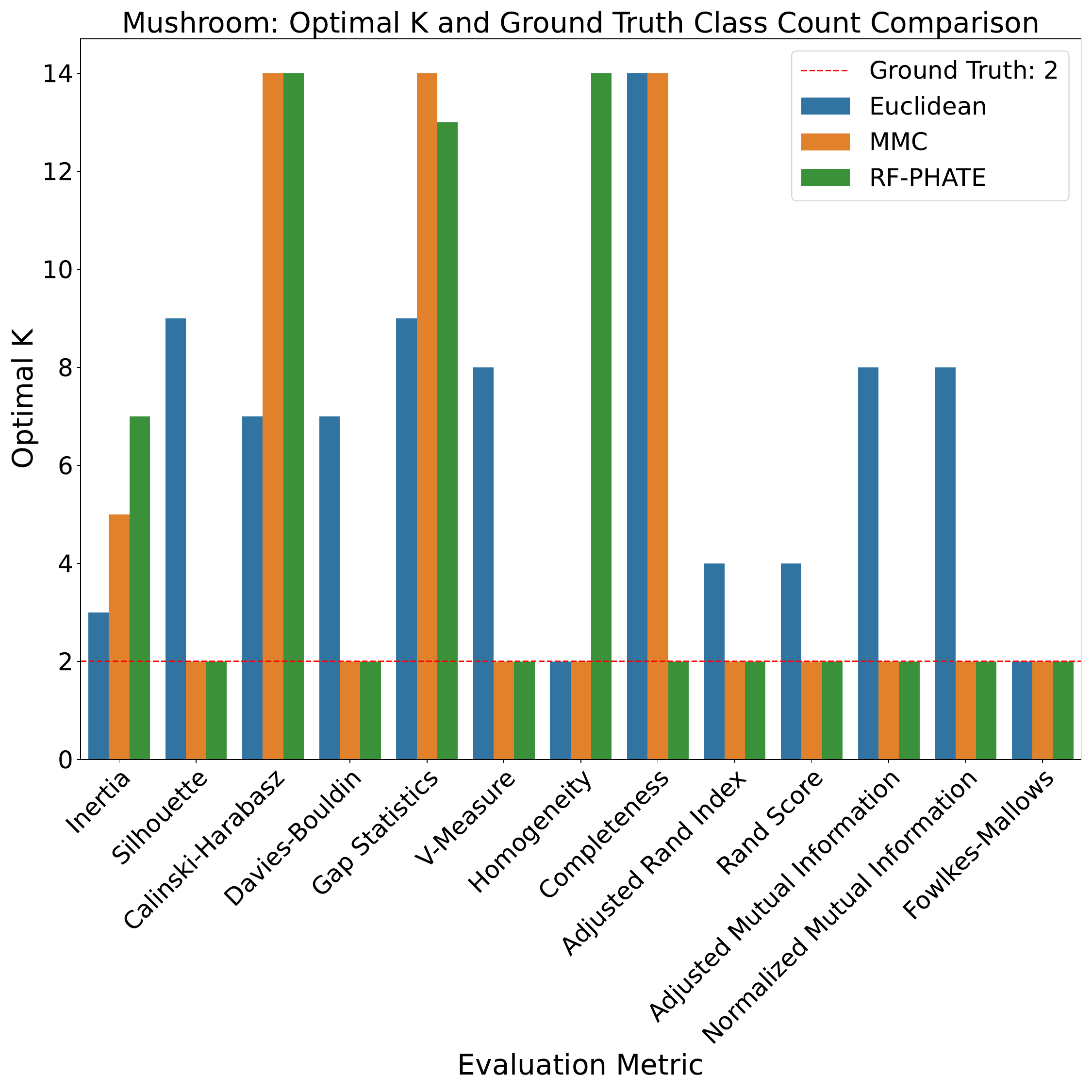}
    \end{subfigure}
    \hfill
    \begin{subfigure}[t]{0.2\textwidth}
        \includegraphics[width=\textwidth]{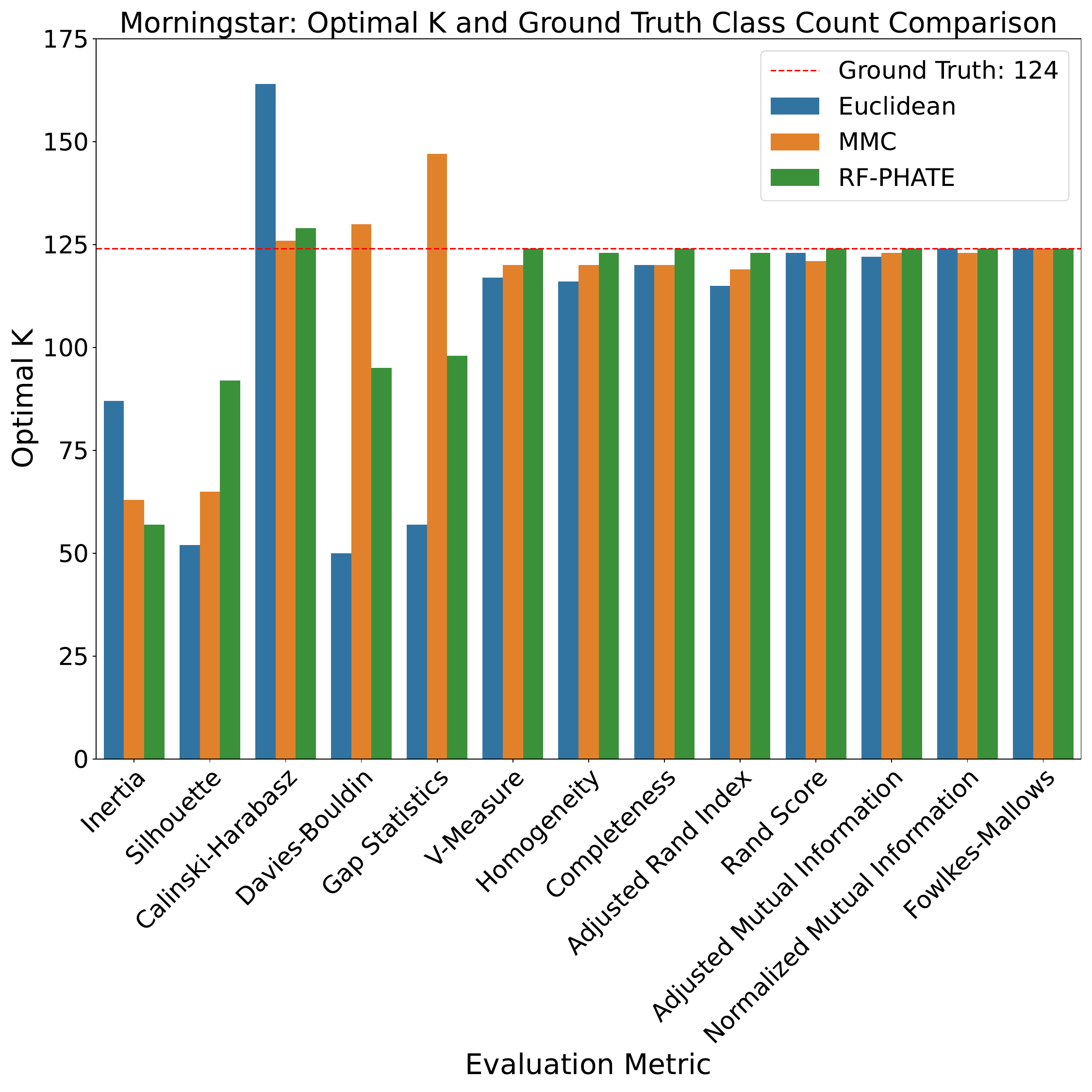}
    \end{subfigure}
\caption{Plots comparing optimal number clusters $K_C$ based on different evaluation metrics and the number of ground truth classes (horizontal line). Further explanation of the figures is provided in Section \ref{sec:results}.} \label{fig:no_clusters_vs_no_of_classes}
\end{figure*}

Figure \ref{fig:no_clusters_vs_no_of_classes} summarizes the results of K-means clustering with all three distance metrics with respect to all the evaluation metrics used in this study: the horizontal line in each figure corresponds to the number of ground truth classes. Each bar corresponds to $K_C$, the optimal number of clusters, as obtained with respect to the specific internal or external evaluation metric for different distance metrics distinguished by different colors on the bars.

For the Euclidean distance, the evaluation metrics demonstrated varied performance. The Silhouette Score and Calinski-Harabasz Index both achieved an exact match ($K_C$ equals the number of ground truth classes) around 50\% of the time. In contrast, the Inertia metric and Davies-Bouldin index showed poor performance rarely reproducing the number of ground truth classes. In general, the internal metrics often fail when attempting to reproduce the ground truth classes as clusters. The external metrics achieved better performance with the Fowlkes-Mallows Score being the best performer, achieving an exact match 70\% 

For the K-means clustering with MMC method, the performance of the internal evaluation metrics showed an improvement over Euclidean clustering. The Silhouette Score achieved an exact match 53\% of the time and a discrepancy of 2 or less in around 64\% of the datasets, whereas Inertia continued to perform poorly with hardly any exact match. The Calinski-Harabasz Index and Davies-Bouldin Index showed moderate performance, but their accuracy was lower than the Silhouette Score. External evaluation metrics also improved their performance, whereas the Fowlkes- Mallows score and Rand score both achieved an exact match 70\% of the time, the Adjusted Rand score, V-Measure, Adjusted Mutual Information score, and Normalized Mutual Information score obtained exact match percentages ranging from around 53\% to 70\%.

Finally, the K-means clustering with RF-PHATE transformations significantly improved the performance of internal evaluation metrics. Inertia showed the most notable improvement, achieving an exact match around 65\% of the time and a discrepancy of 2 or less in 82\% of the datasets. Similarly, Silhouette score and Davies-Bouldin index also improved with an exact match around 60\% and 65\% times respectively. As for the external metrics, the Fowlkes-Mallows Score and Rand Score both achieved an exact match around 75\% of the time, with the Rand Score showing a discrepancy of 2 or less 100\% of the time and the Fowlkes-Mallows Score 94\% of the time. The Adjusted Rand score, Adjusted Mutual Information score, and Normalized Mutual Information score showed the highest exact match. 

The results for the Morningstar dataset follow a similar pattern: For the Euclidean k-means clustering, internal evaluation metrics such as Inertia and the Silhouette score suggested $K_C$ of 57 and 52 respectively, which is significantly lower than the actual number of categories (124). The MMC and RF-PHATE based clustering demonstrated increasingly better performances in accurately capturing the data structure, respectively, with obtaining the perfect match with the ground truth class count for six out of the eight external evaluation metrics, with the remaining two metrics being only one-off from the ground truth class count, i.e., $K_C = 123$.

The subsequent improvement in reproducing the number of ground truth classes with the use of better and more nuanced distance metric algorithms supports the hypothesis that the use of a proper distance metric for clustering is critical if the goal of clustering is to reproduce the ground truth classes. Moreover, the superior performance of RF-PHATE suggests that advanced clustering techniques can uncover relevant patterns and highlight the importance of using sophisticated methods to detect subtle and meaningful structures within complex datasets.

\subsection{Visualization of Clusters using MDS Plots}
\begin{figure*}[htbp]
    \centering
    \begin{subfigure}[t]{0.49\textwidth}
        \raisebox{-\height}{\includegraphics[width=\textwidth]{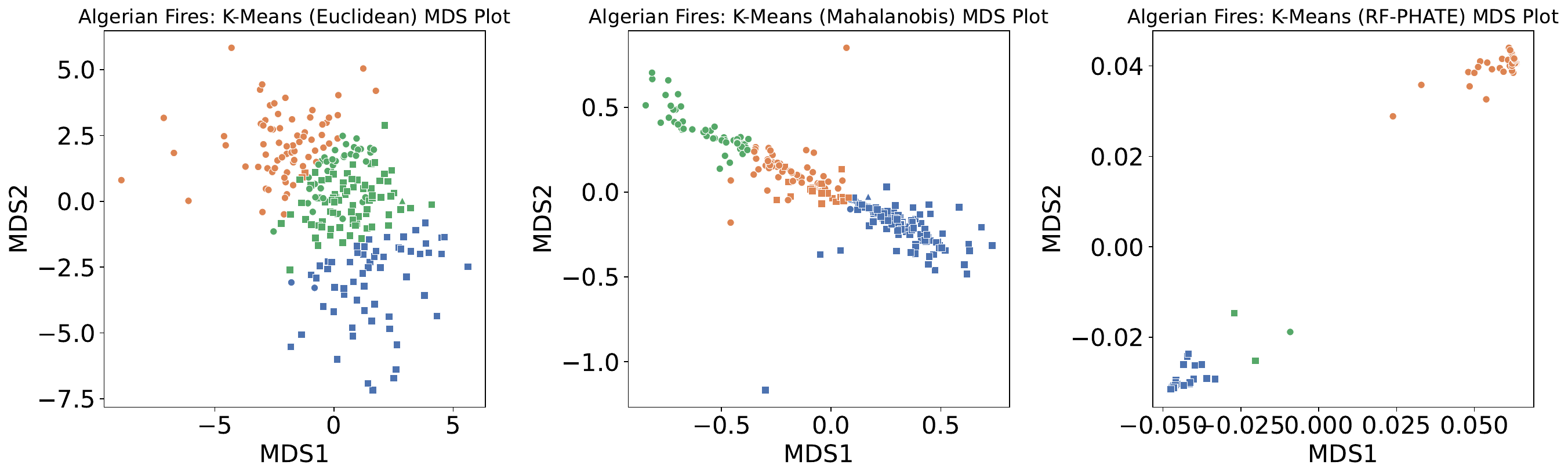}}
    \end{subfigure}
    \hfill
    \begin{subfigure}[t]{0.49\textwidth}
        \raisebox{-\height}{\includegraphics[width=\textwidth]{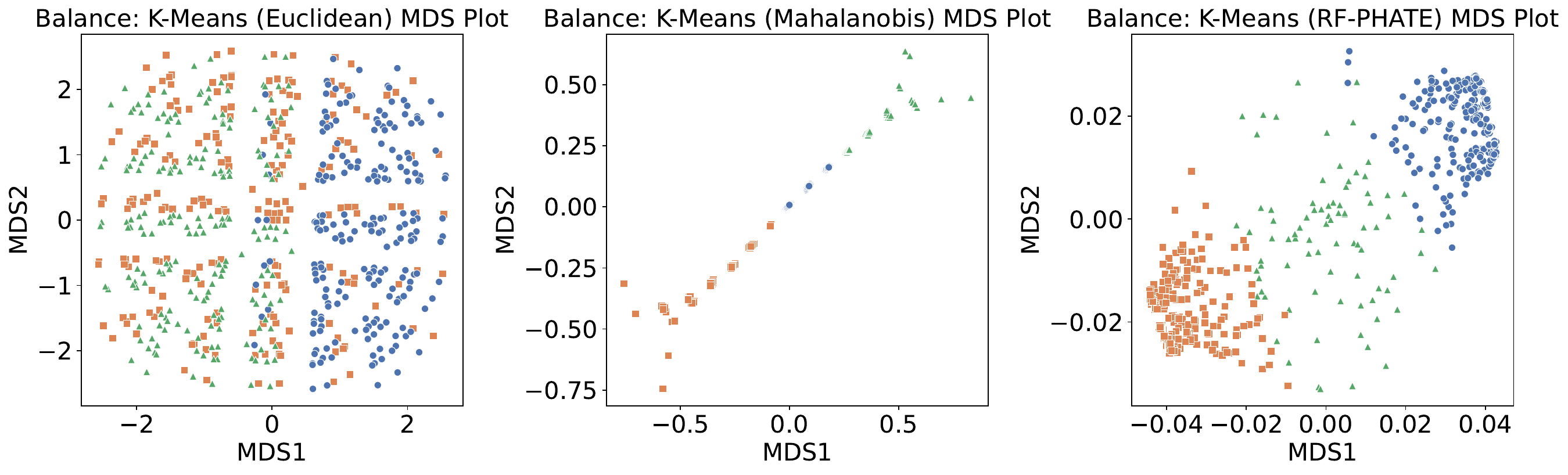}}
    \end{subfigure}

    \begin{subfigure}[t]{0.49\textwidth}
        \raisebox{-\height}{\includegraphics[width=\textwidth]{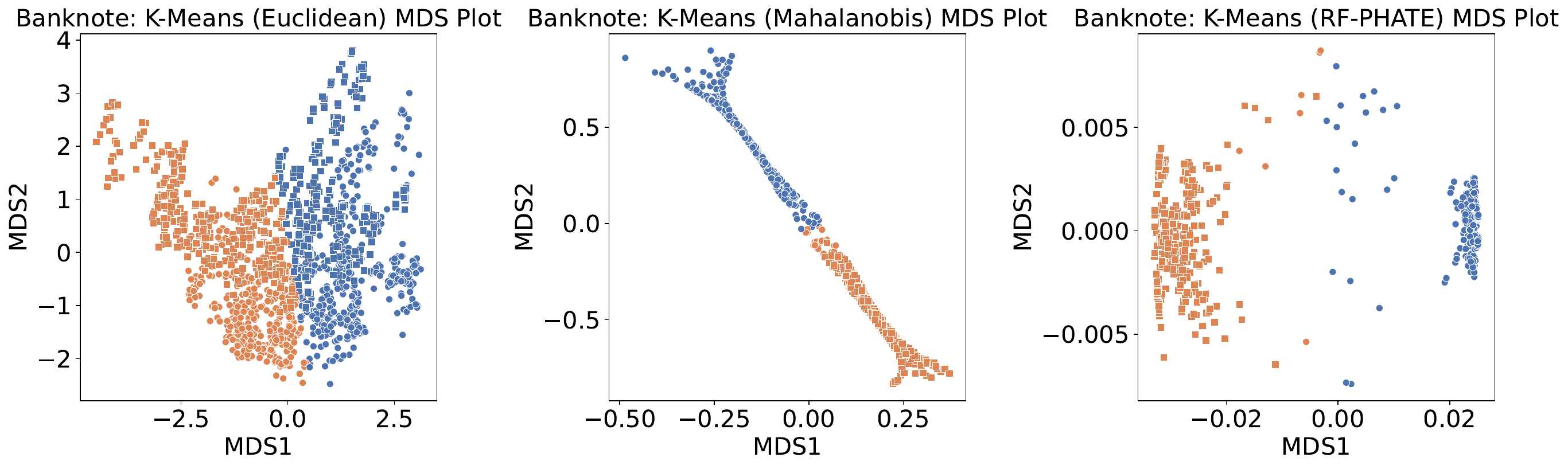}}
    \end{subfigure}
    \hfill
    \begin{subfigure}[t]{0.49\textwidth}
        \raisebox{-\height}{\includegraphics[width=\textwidth]{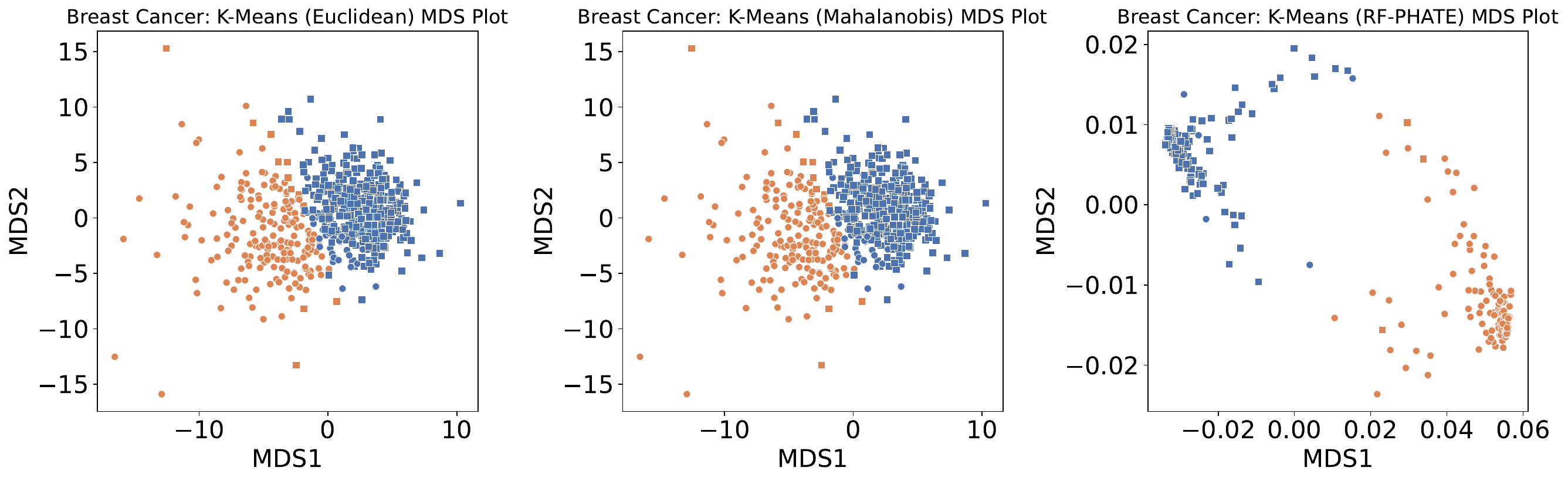}}
    \end{subfigure}

    \begin{subfigure}[t]{0.49\textwidth}
        \raisebox{-\height}{\includegraphics[width=\textwidth]{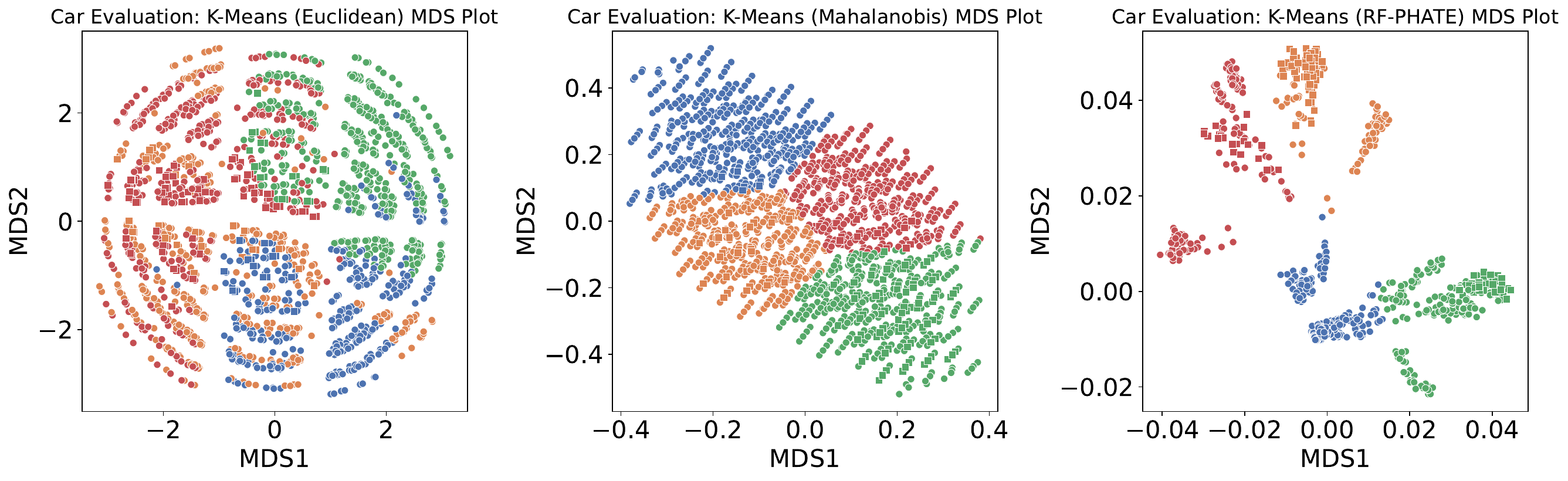}}
    \end{subfigure}
    \hfill
    \begin{subfigure}[t]{0.49\textwidth}
        \raisebox{-\height}{\includegraphics[width=\textwidth]{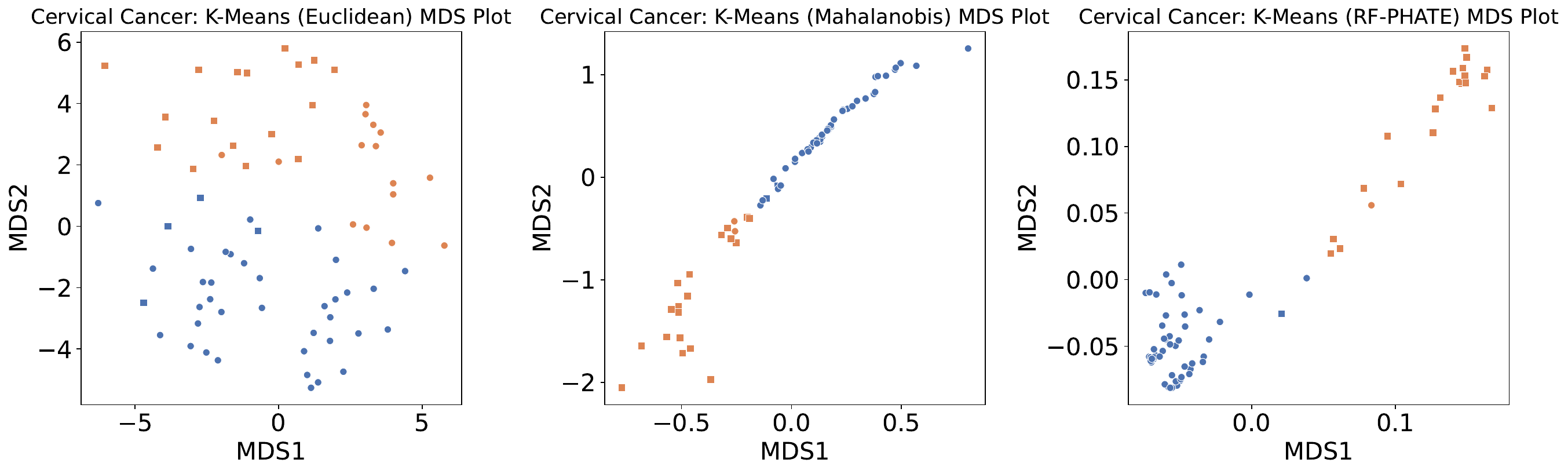}}
    \end{subfigure}

    \begin{subfigure}[t]{0.49\textwidth}
        \raisebox{-\height}{\includegraphics[width=\textwidth]{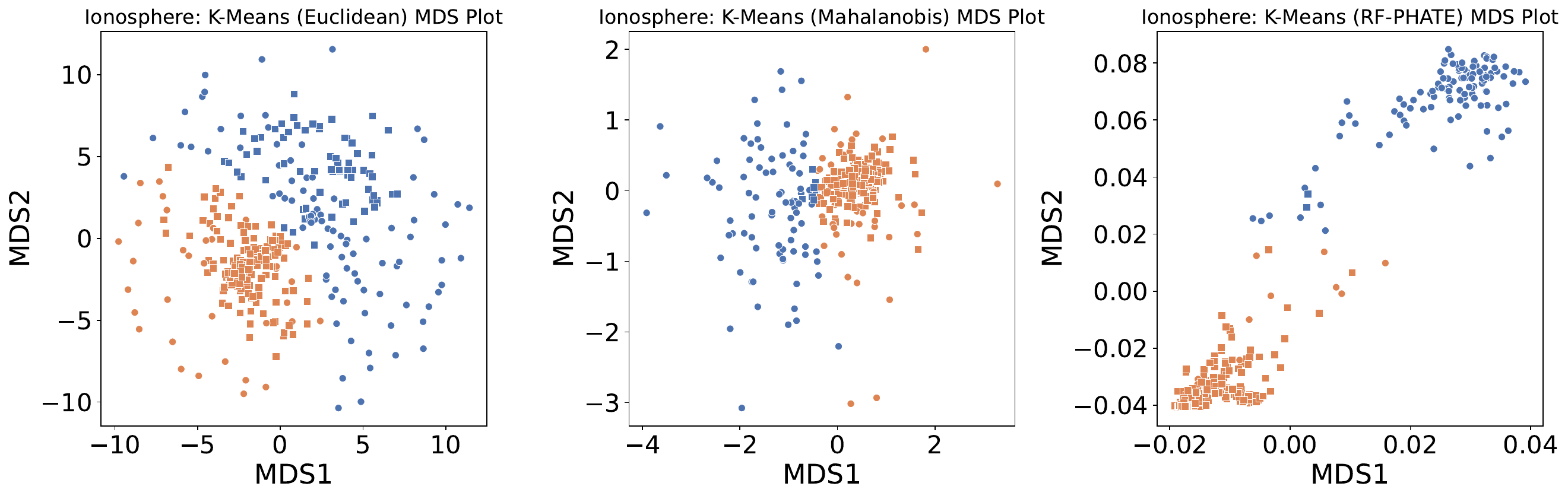}}
    \end{subfigure}
    \hfill
    \begin{subfigure}[t]{0.49\textwidth}
        \raisebox{-\height}{\includegraphics[width=\textwidth]{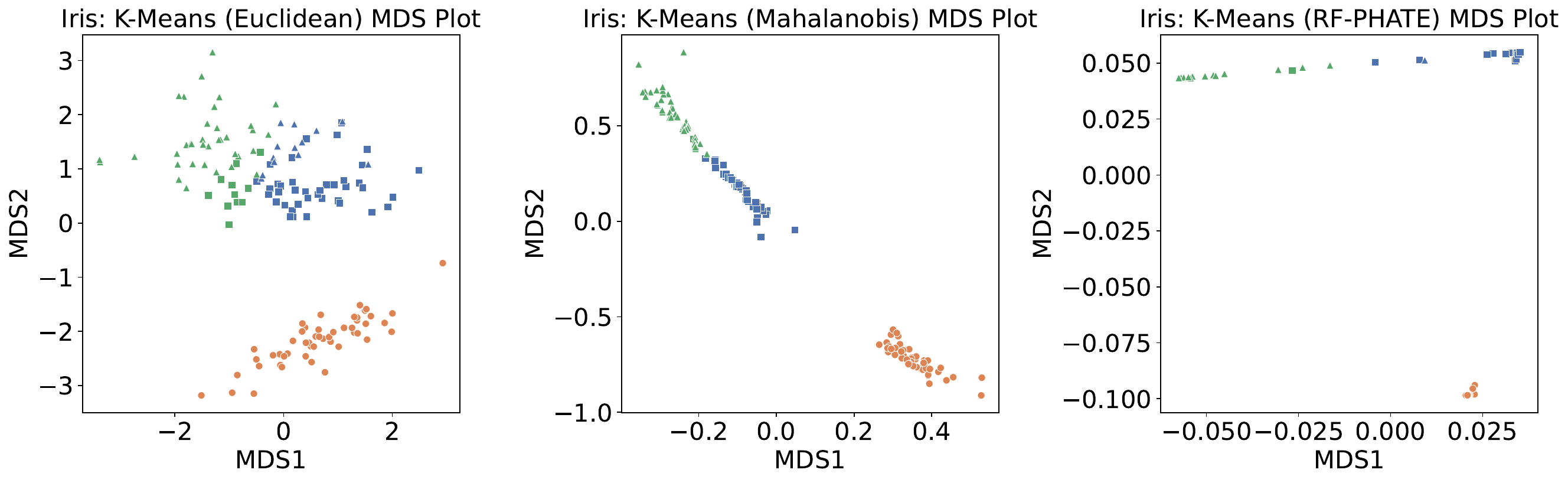}}
    \end{subfigure}

    \begin{subfigure}[t]{0.49\textwidth}
        \raisebox{-\height}{\includegraphics[width=\textwidth]{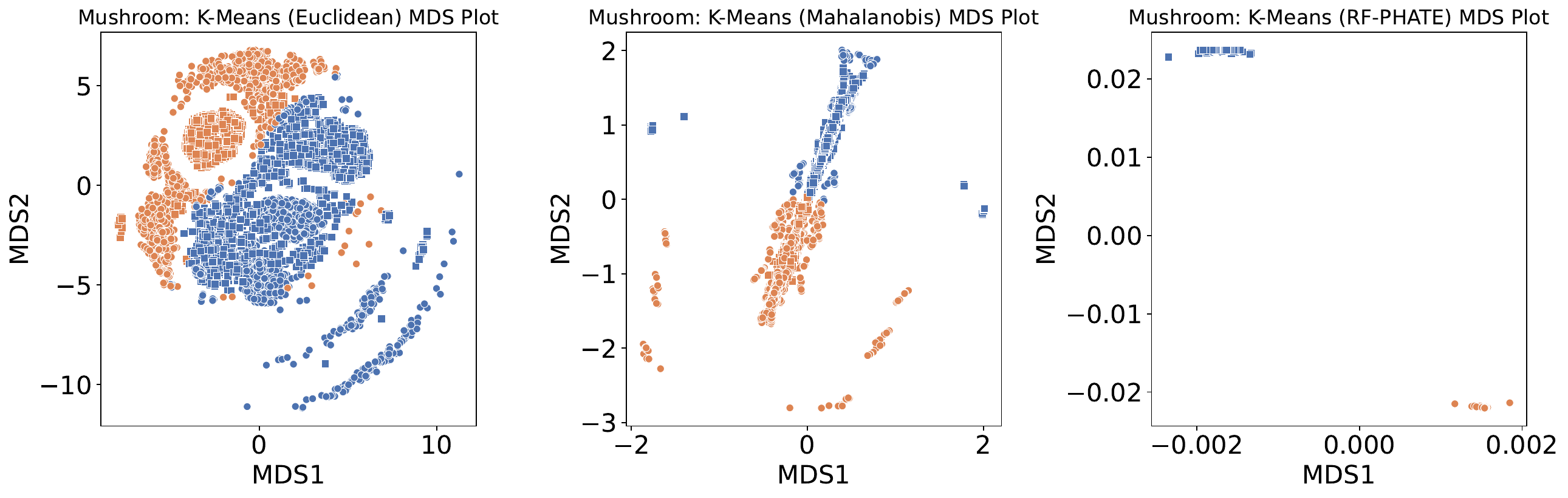}}
    \end{subfigure}
    \hfill
    \begin{subfigure}[t]{0.49\textwidth}
        \raisebox{-\height}{\includegraphics[width=\textwidth]{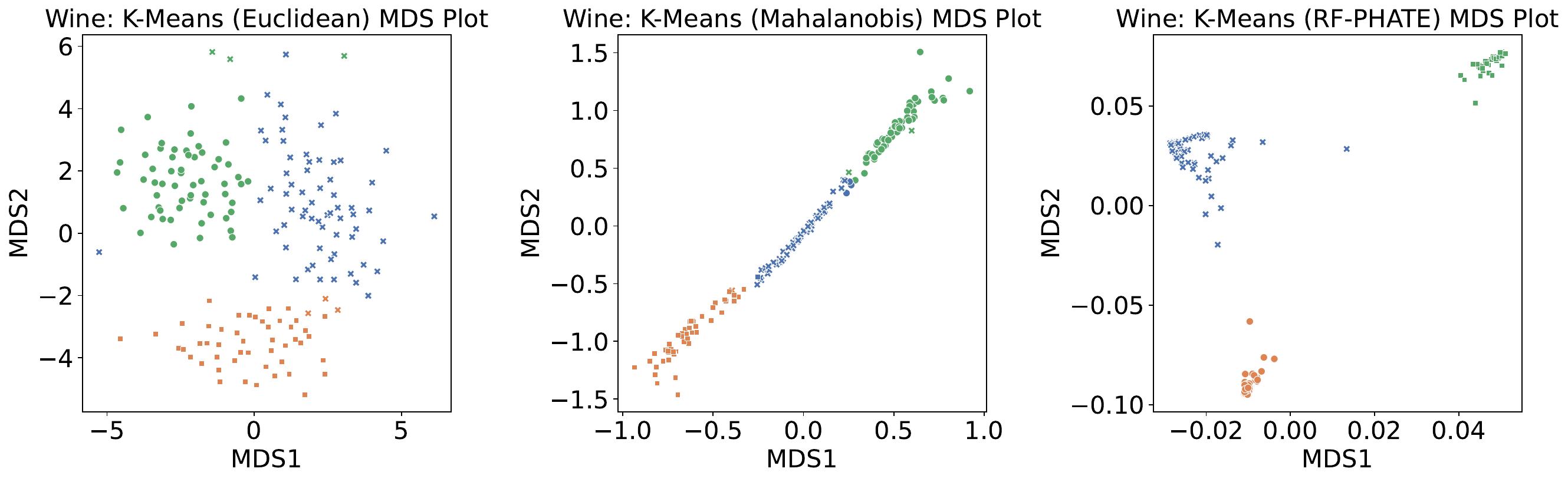}}
    \end{subfigure}

    \begin{subfigure}[t]{0.33\textwidth}
        \includegraphics[width=1.2\textwidth,trim= 0 0 7.5mm 0 ,clip=true]{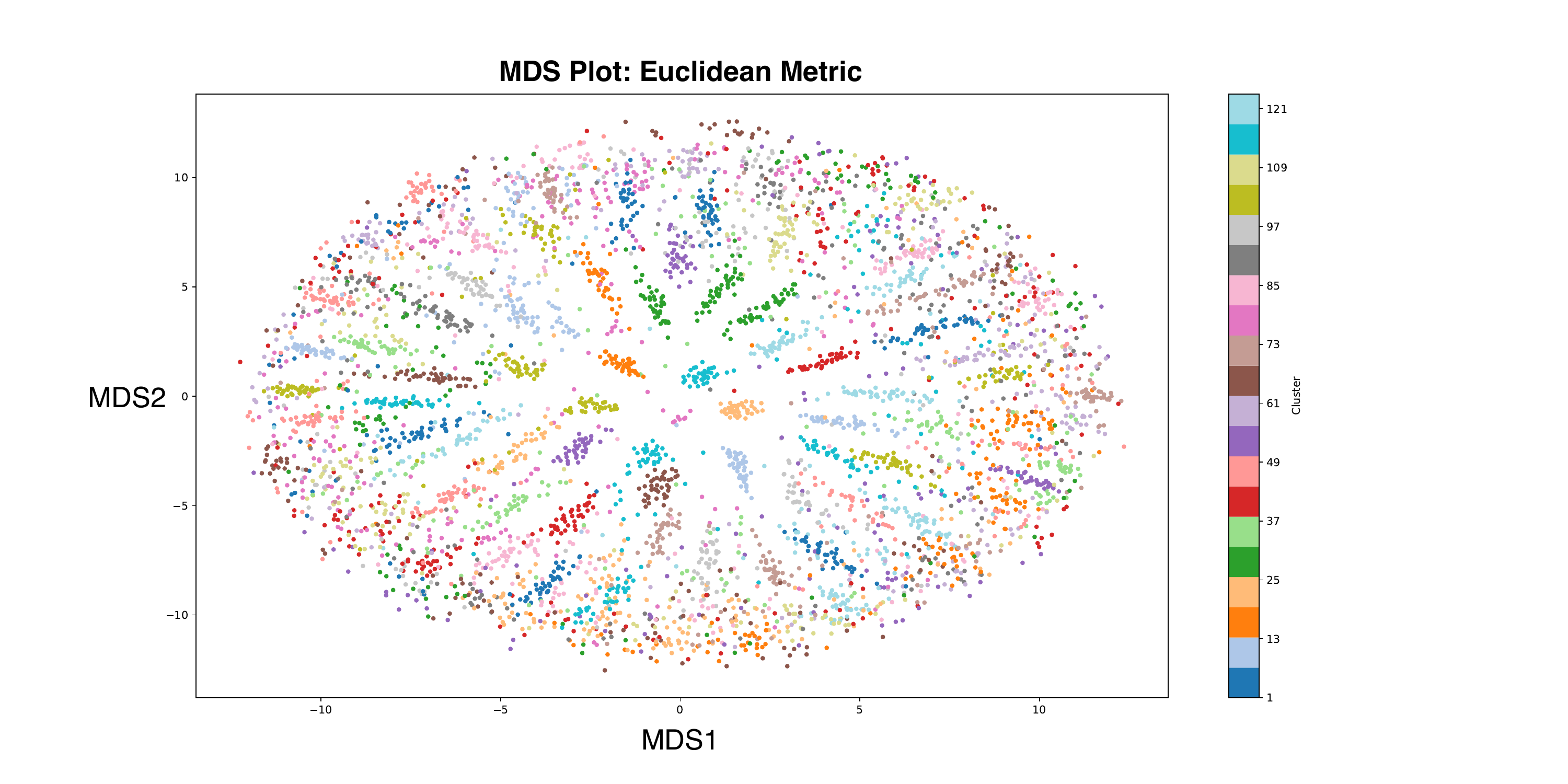}
    \end{subfigure}
    \hfill
    \begin{subfigure}[t]{0.33\textwidth}
        \includegraphics[width=1.2\textwidth, trim= 0 0 7.5mm 0 ,clip=true]{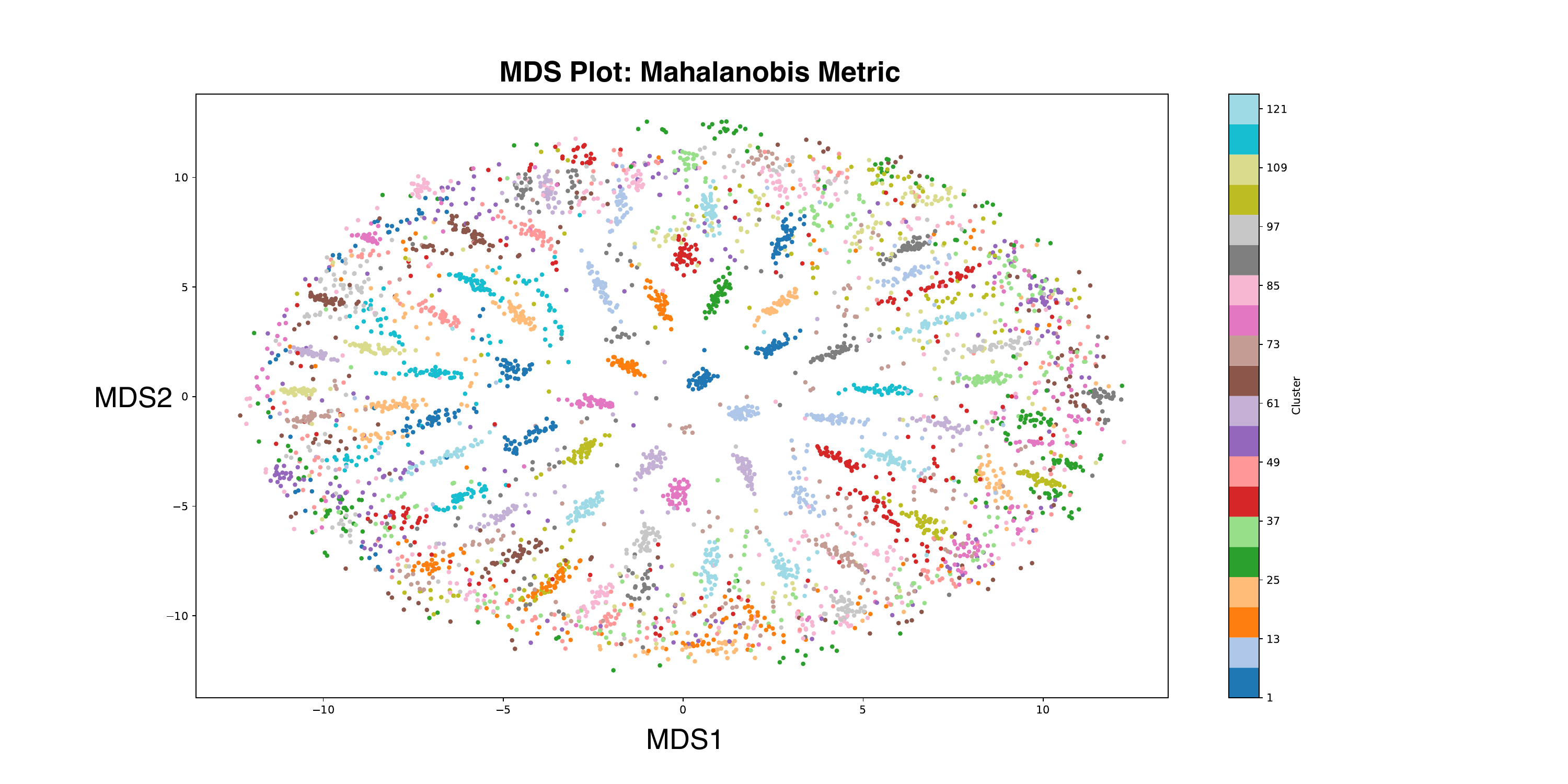}
    \end{subfigure}
    \hfill
    \begin{subfigure}[t]{0.33\textwidth}
        \includegraphics[width=1.2\textwidth,trim= 0 0 7.5mm 0, clip=true]{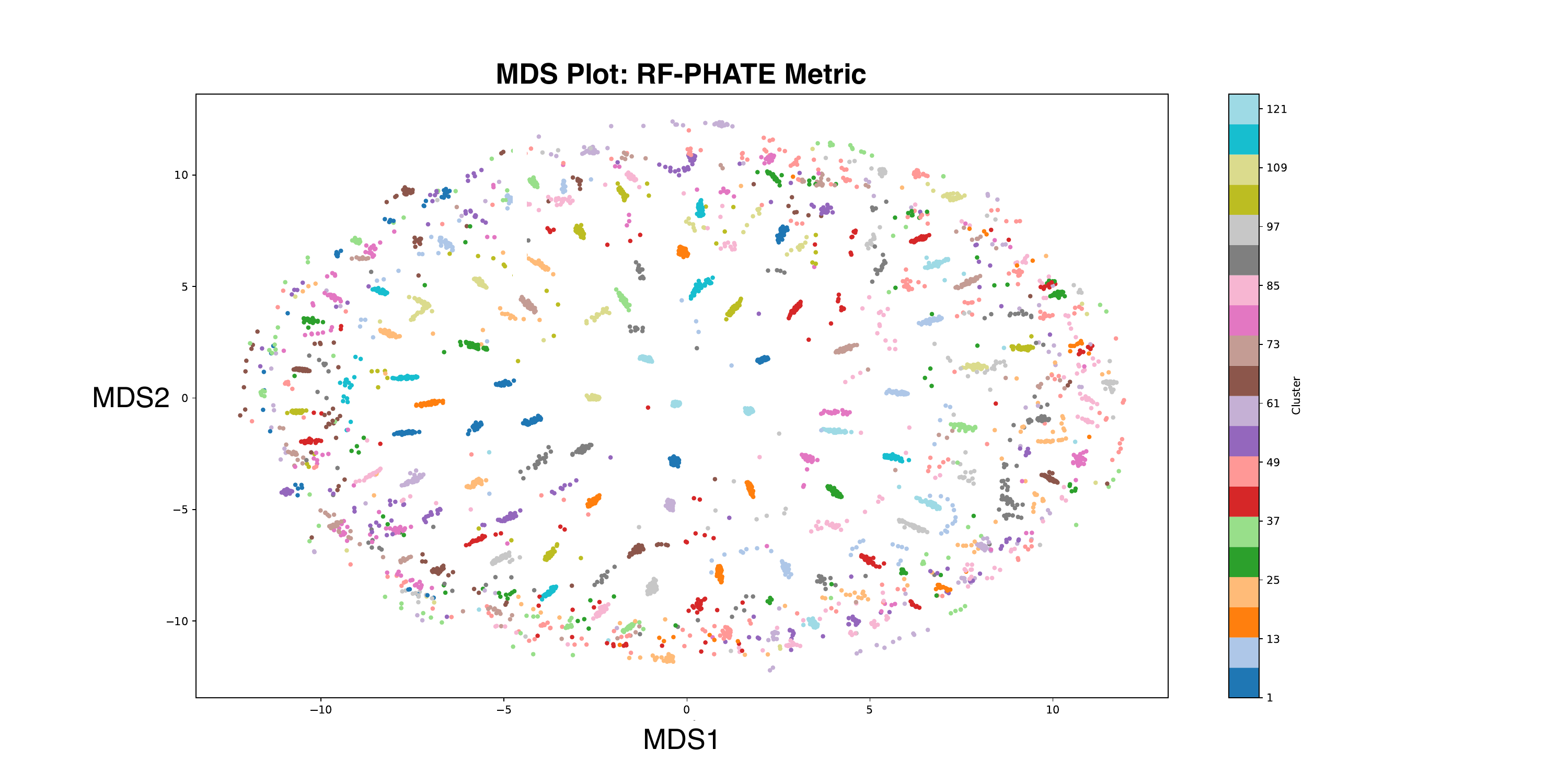}
    \end{subfigure}
\caption{MDS Plots for Public datasets and Funds data using three different distance metrics. The colors represent different ground-truth classes.} \label{fig:MDS_plots}
\end{figure*}

In Figures \ref{fig:MDS_plots}, we visualize each of the datasets including the fund's data with two components of MDS. Although a 2D slice is not a complete representation of high dimensional data, it is intriguing to visually inspect the effect of various distance metrics on the clustering where the RF-PHATE, though not surprisingly, most cleanly separates data points of different classes in distinct clusters. 

\subsection{Correlation Analysis}
To investigate the effect of available features and their predictive power for the target variable on reproducing the ground truth classes, we compute correlations between the weighted F1 score
for each of the internal and external clustering evaluation metrics, for each of the datasets. Here, we hypothesize that a clustering algorithm, even with an appropriate distance metric,  can only be expected to reproduce the ground truth classes as distinct clusters if the RF model can accurately learn the ground truth classes using the available input features in the first place.
\begin{table}[ht!]
\centering
\begin{tabular}{l l l l}
\toprule
Dataset         &       Euclidean & Mahalanobis &      RF-PHATE \\
\midrule
Mushroom        &            0.60 &        0.86 &          0.97 \\
Car evaluation  &            0.49 &        0.50 &          0.91 \\
Iris            &            0.72 &        0.97 &          0.98 \\
Wine            &            0.95 &        0.93 &          1.00 \\
Breast cancer   &            0.83 &        0.86 &          0.92 \\
Ionosphere      &            0.58 &        0.62 &          0.93 \\
Balance         &            0.57 &        0.71 &          0.94 \\
Algeria fires   &            0.62 &        0.71 &          0.97 \\
Banknote        &            0.50 &        0.95 &          0.98 \\
Cervical cancer &            0.60 &        0.91 &          0.94 \\
\bottomrule
\end{tabular}
\caption{Clustering accuracy (Eq.~\ref{eq:accuracy}) using different methods.}
\label{tab:comparison_clustering}
\vspace{-6mm}
\end{table}
In Tables \ref{tab:rf_correlations} and \ref{tab:comparison_clustering}, we calculate correlations between the weighted F1 score and each of the internal and external clustering evaluation metrics for each of the three methods to determine a relationship between the predictive power of the input features and the evaluation metrics. We find that the RF’s performance is strongly correlated with the clustering performance across various evaluation metrics, especially the external clustering metrics, such as V-measure, Adjusted Rand index, and Fowkles-Mallows Score.

\begin{table}[ht!]
    \centering
    \small 
    \begin{tabular}{lccc}
        \toprule
        \textbf{Metric} & \textbf{Euclidean} & \textbf{Mahalanobis} & \textbf{RF-PHATE} \\
        \midrule
        Inertia & 0.30 (±18855.5) & 0.32 (±360.8) & -0.26 (±0.1) \\
        Silhouette & -0.08 (±0.09) & 0.35 (±0.12) & 0.60 (±0.16) \\
        Calinski-Harabasz & -0.03 (±1388.6) & 0.38 (±3895.6) & 0.30 (±52582.2) \\
        Davies-Bouldin & 0.21 (±0.36) & -0.19 (±0.24) & -0.46 (±0.25) \\
        \midrule
        V-Measure & 0.71 (±0.20) & 0.83 (±0.23) & 0.81 (±0.29) \\
        Homogeneity & 0.63 (±0.20) & 0.73 (±0.23) & 0.83 (±0.34) \\
        Completeness & 0.75 (±0.26) & 0.84 (±0.31) & 0.74 (±0.29) \\
        Adj. Rand Index & 0.67 (±0.16) & 0.74 (±0.21) & 0.73 (±0.31) \\
        Rand Index & 0.67 (±0.16) & 0.74 (±0.21) & 0.70 (±0.17) \\
        Adj. Mutual Info & 0.73 (±0.20) & 0.84 (±0.23) & 0.82 (±0.30) \\
        Norm. Mutual Info & 0.71 (±0.20) & 0.83 (±0.23) & 0.81 (±0.29) \\
        Fowlkes-Mallows & 0.57 (±0.13) & 0.65 (±0.17) & 0.70 (±0.22) \\
        \midrule
        Clustering Accuracy & 0.61 (±0.14) & 0.84 (±0.16) & 0.91 (±0.17) \\
        \bottomrule
    \end{tabular}
    \caption{Correlation between RF performance and clustering metrics. The values in parentheses indicate the standard deviation of the corresponding metric.}
    \label{tab:rf_correlations}
    \vspace{-6mm}
\end{table}

Table \ref{tab:rf_correlations} shows that relatively weaker correlations were observed with several external clustering metrics such as V-Measure, Completeness, Adjusted Rand Index, Adjusted Mutual Information, and Normalized Mutual Information, indicating that higher RF performance for a dataset is strongly associated with improved clustering quality and performance. For the Mahalanobis and RF-PHATE distance metrics, the correlation is even stronger, which is not surprising as both the external metrics and supervised distance metrics learning methods are directly tied to the ground truth labels.

\section{Conclusion}
Many financial and other industry datasets come with a manual categorization system (e.g., Morningstar and Lipper categorizations of funds, fund and bond ratings, GICS classification, etc.) where experts assign unique labels to each data point based on the available data and their domain expertise, though the decision-making process is often not transparent to end users. Machine learning techniques can naturally be employed to approximate this decision-making process. However, researchers often use unsupervised clustering algorithms on the available variables (while masking the target variable) to reproduce the categorization system and may question its validity if it fails. This issue extends beyond the financial domain: when a novel unsupervised clustering algorithm is proposed, its effectiveness is demonstrated using publicly available labeled datasets, with overlap measured by metrics like V-measure between the clusters from the algorithm and ground truth classes.

We argued that unsupervised clustering algorithms with arbitrarily chosen variables and distance metric may be inappropriate and lead to incorrect conclusions: the labels arise from a categorization system where the experts committee (or any other system to label the data points) may assign labels based on specific pre-chosen variables and a specific distance metric. 

With extensive experiments on various publicly available toy datasets as well as mutual funds dataset with Morningstar categorization, and using multiple internal and external evaluation metrics for clustering algorithms, we showed that if the available variables are predictive enough for the ground truth labels and the distance metric is learned using the ground truth labels with supervised distance metric learning methods such as the Mahalanobis distance metric learning \cite{dml-xing} or RF-PHATE based technique \cite{moon2019visualizing, rf-phate-rhodes}, then even a simple clustering algorithm such as K-means can accurately reproduce the ground truth classes. We measured the predictive power of the variables by computing the accuracy of the RF to learn the ground truth labels using the available input features. We also showed that RF-PHATE based distance metric learning technique has various advantages over the traditional Mahalanobis metric-based technique in that the former can be employed on mixed variable type datasets, can handle missing values, scales well with the size of the dataset, requires minimal preprocessing, and learns local (adaptive to the nuances of different data regions) data structures, and outperforms the latter method as measured by most of the internal and external evaluation metrics for clustering.

Furthermore, we also investigate the effectiveness of the clustering evaluation metrics and find that the Fowlkes-Mallows Score and Rand Score, consistently provide more reliable assessments of clustering quality strictly evaluated against the ground truth labels. In contrast, internal metrics like the Silhouette Score and Calinski-Harabasz Index, while useful, are far less likely to provide an optimal \( k \) that aligns with the ground-truth class count. 

In one sense, it should come as no surprise that clustering with a distance metric that is trained using certain ground truth class labels would demonstrate superior performance with respect to external evaluation metrics that are based on those same class labels. What was not obvious to us is that it should be possible to recover such a classification system using simple unsupervised clustering methods on a feature space of mixed data types that does not include these labels. What our analysis reveals, in short, is the power of supervised metric learning to encode the categorical structure of a given classification system in an independent set of features, provided these features contain enough information to reliably predict these classes. The consistency of a categorization system such as Morningstar, then, is more fairly assessed not based on whether it coincides with clusters obtained through just \textit{any} distance metric on some previously agreed-upon feature space, but based on whether it coincides with clusters obtained through \textit{some} distance metric on that space - a condition that is not trivial to satisfy, particularly as the features may lack sufficient information to predict these categories.

Finally, our conclusion about the effectiveness of the internal evaluation metrics for unsupervised clustering closely matches with that of Ref.~\cite{ma2023need} where they empirically investigated internal evaluation strategies for hyperparameters of unsupervised models for anomaly detection problems with respect to random selection of hyperparameters and the popular state-of-the-art detector Isolation Forest with default hyperparameters. They concluded that none of the existing and adapted strategies for hyperparameter selection for unsupervised anomaly detection models would be significantly different from the simple baseline method. Our findings indicate the same for hyperparameter tuning (in this case, of K, the optimal number of clusters) for a completely unsupervised K-means algorithm (i.e., arbitrarily selected distance metric such as the Euclidean distance). Our study demands a further thorough investigation of the effectiveness of the internal evaluation metrics with respect to other clustering algorithm methods than the K-means algorithm.

\section{Acknowledgement}
The views expressed here are those of the authors alone and not of BlackRock, Inc. We are grateful to Joshua Rosaler, Guanchao Feng, Jake Rhodes, Kevin Moon and Kelvyn Bladen for their valuable inputs to this work.

\bibliographystyle{unsrt}
\bibliography{sample-base}
\end{document}